\documentclass[letterpaper, 10 pt, conference]{ieeeconf}  

\IEEEoverridecommandlockouts                              
                                                          
\usepackage{url}
\usepackage{comment}
\usepackage{amsmath,amssymb,amsfonts, mathtools}
\usepackage{graphicx}
\usepackage{bm}
\usepackage{epstopdf}
\usepackage{placeins}
\usepackage{bm}
\usepackage{cite}
\let\labelindent\relax
\usepackage[inline]{enumitem}
\usepackage{multicol}
\usepackage{tikz}
\usepackage{caption}
\usepackage{subcaption}
\usepackage{soul}
\usepackage{wrapfig}
\usepackage{booktabs}
\usepackage[normalem]{ulem}
\useunder{\uline}{\ul}{}
\usepackage{comment}
\usepackage{xcolor}
\usepackage[linesnumbered,ruled,vlined]{algorithm2e}
\SetKwInput{KwInput}{Input}                
\SetKwInput{KwOutput}{Output}              

\usepackage[inline]{enumitem}

\SetKwRepeat{Do}{do}{while}%

\newcommand{\vect}[1]{\boldsymbol{#1}}

\DeclareMathOperator*{\argmin}{arg\,min}
\definecolor{bblue}{RGB}{31,119,180}
\title{\LARGE \bf Robot Learning from Demonstration Using Elastic Maps}

\author{Brendan Hertel*, Matthew Pelland*, and S. Reza Ahmadzadeh
	\thanks{Persistent Autonomy and Robot Learning (PeARL) Lab, University of Massachusetts Lowell, MA, 01854. Email: \tt\small\{brendan\_hertel, matthew\_pelland\}@student.uml.edu,reza@cs.uml.edu}
	\thanks{* Indicates equal contribution.}}

\begin{document}

\maketitle
\thispagestyle{empty}
\pagestyle{empty}

\begin{abstract}
Learning from Demonstration (LfD) is a popular method of reproducing and generalizing robot skills from human-provided demonstrations. In this paper, we propose a novel optimization-based LfD method that encodes demonstrations as \emph{elastic maps}. An elastic map is a graph of nodes connected through a mesh of springs. We build a skill model by fitting an elastic map to the set of demonstrations. The formulated optimization problem in our approach includes three objectives with natural and physical interpretations. The main term rewards the mean squared error in the Cartesian coordinate. The second term penalizes the non-equidistant distribution of points resulting in the optimum total length of the trajectory. The third term rewards smoothness while penalizing nonlinearity. These quadratic objectives form a convex problem that can be solved efficiently with local optimizers. We examine nine methods for constructing and weighting the elastic maps and study their performance in robotic tasks. We also evaluate the proposed method in several simulated and real-world experiments using a UR5e manipulator arm, and compare it to other LfD approaches to demonstrate its benefits and flexibility across a variety of metrics.
\end{abstract}

\section{Introduction}
\label{sec:intro}

As robots are becoming increasingly present in the world, equipping them with methods that enable humans to teach new skills effortlessly and robustly has become prevalent. One of the most effective methods to teach robots new skills is Learning from Demonstration (LfD). LfD methods enable robots to model a skill demonstrated by a human teacher and use it to reproduce and generalize the skill to novel situations. Existing LfD approaches use various methods for representing and interpreting the demonstrated trajectories such as dynamical systems~\cite{pastorDMP2009}, statistical and probabilistic frameworks~\cite{Calinon2007GMM, Paraschos2013ProMP}, or geometric models~\cite{Ahmadzadeh2018TLGC}. Regardless of the type of representation, many LfD approaches formalize an optimization problem for the adaptation of the demonstrations to novel situations. 

\begin{figure}[t]
\centering
\includegraphics[trim=0 0em 0 0, clip, width=0.41\columnwidth]{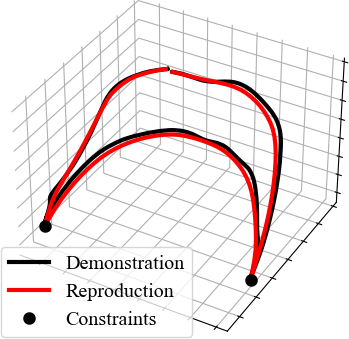}
\includegraphics[trim=0 0em 0 0, clip, width=0.54\columnwidth]{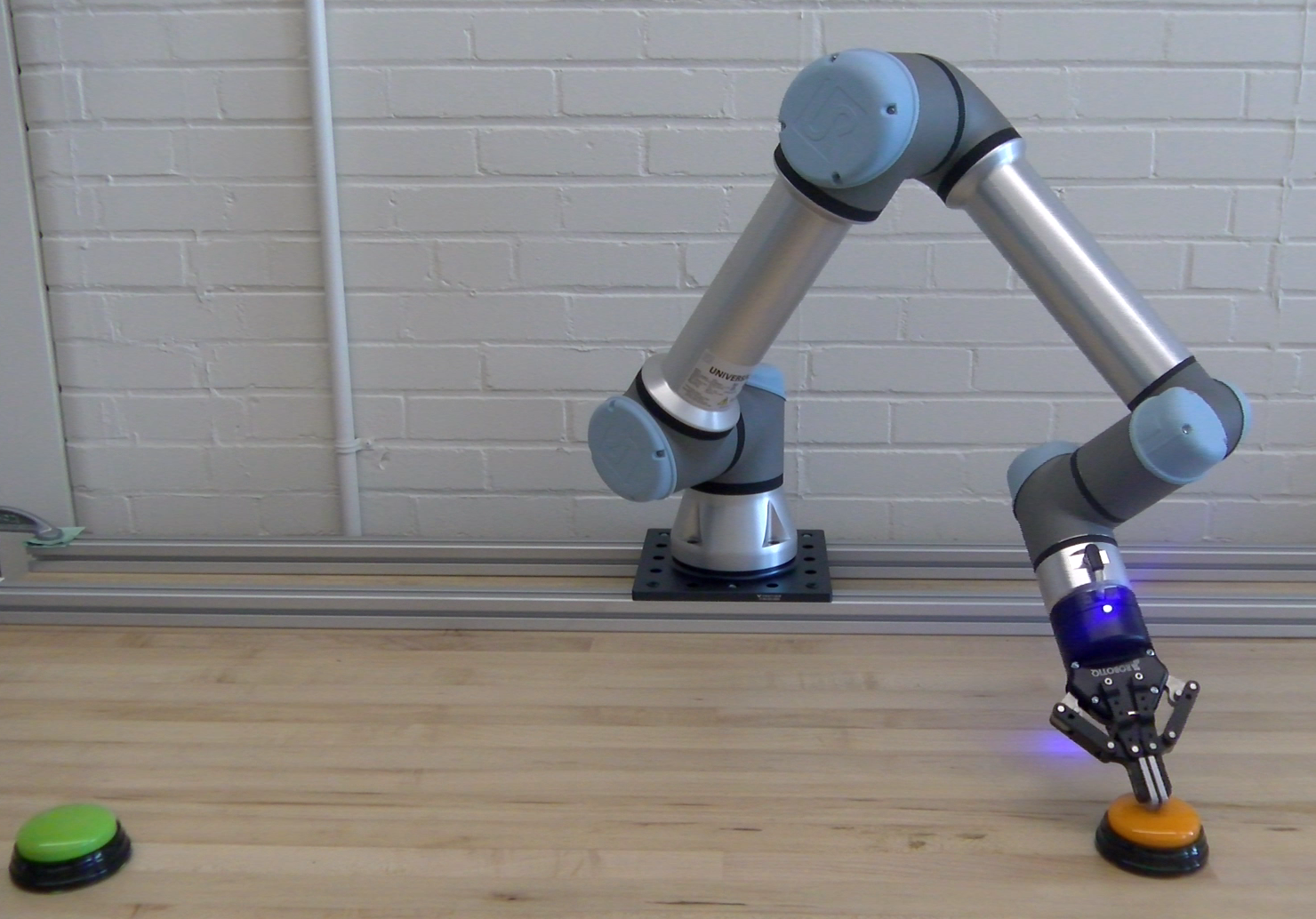}
\caption{\small{An elastic map reproduction of the pressing skill on a UR5e robot. The pressing points are constrained as via-points.}} \label{fig1_trajs}
\end{figure}

Despite promising features, existing LfD representations suffer from shortcomings. 
The generalization ability of a representation can be largely task-dependent, meaning that no method is perfectly suited to every task~\cite{hertel2021SAMLfD}. To resolve this, some representations incorporate parameters which can be tuned for each task. However, these parameters must be tuned by the user, and can be difficult for inexperienced users to tune properly. Furthermore, most LfD representations can incorporate either single~\cite{pastorDMP2009} or multiple~\cite{Calinon2007GMM, Ahmadzadeh2018TLGC, calinon2010learning_CorrDMP} demonstrations, but not both. While providing single demonstrations might be more convenient for the user, multiple demonstrations provide more information about the skill and environment. 
Another issue with many LfD representations is failure to incorporate different types of constraints, particularly via-point constraints~\cite{pastorDMP2009}. These constraints may be necessary for the successful reproduction of skills such as pressing multiple buttons (shown in Fig.~\ref{fig1_trajs}) or pick-and-place.



To address these issues, we propose a novel LfD approach that encodes a demonstrated skill as an \emph{elastic map} and adapts to novel situations. An elastic map is a graph of nodes in data space connected through a mesh of springs ~\cite{gorban_zinovyev}. Our approach constructs and optimizes skill models using elastic maps. We argue that this approach presents several advantages over previous methods. Our LfD approach is capable of modeling both single and multiple demonstrations, and generates reproductions with desirable features such as smoothness and shape preservation. The objectives have natural interpretations, making parameter tuning potentially more intuitive for inexperienced users. We formulate the elastic map and associated energies to form a convex optimization problem that can be solved efficiently. Additionally, our LfD approach can incorporate any number of initial, final, or via-point constraints smoothly into the reproductions.


We evaluate the proposed method in several simulated and real-world experiments using a UR5e manipulator arm (shown in Fig.~\ref{fig1_trajs}).  We investigate methods for constructing the elastic map from demonstrations and determining elastic map parameters which are best-suited for skill reproductions. To demonstrate its benefits and flexibility across a variety of metrics, we compare our approach to three LfD algorithms. The results show the proposed LfD approach can construct skill models from single or multiple demonstrations, preserves key features such as curvature and smoothness, and can satisfy initial, final, or via-point constraints.



\section{Related Work} 
\label{sec:RW}


Many LfD techniques rely on optimization to reproduce the desired skill by adapting the given demonstrations to novel situations~\cite{Meirovitch2016JA, zucker2013chomp, Ravichandar2019MCCB, hertel2021TLFSD}. These approaches use the data provided in demonstrations to construct a cost function, and optimize this cost function to find a reproduction. Different techniques construct this cost function in different ways. For example, Trajectory Learning from Failed and Successful Demonstrations (TLFSD)~\cite{hertel2021TLFSD} learns from both successful and failed demonstrations of a skill to construct a cost function which rewards success and penalizes failure. Alternatively, Multi-Coordinate Cost Balancing (MCCB)~\cite{Ravichandar2019MCCB} finds important features in demonstrations by encoding them into multiple differential coordinates, and finding a reproduction which best models these features. Although multiple demonstrations provide more information to the encoding process, some LfD techniques use data from a single demonstration, such as the Jerk-Accuracy Model~\cite{Meirovitch2016JA}. This approach finds an optimal reproduction by trading off between convergence to the demonstrated trajectory and smoothness. LfD techniques which use optimization to find a reproduction can be slow if the constructed cost function is non-convex. Additionally, these optimization-based techniques construct separate objective terms for each feature, and usually balance these terms through user-defined parameters.


One of the most common LfD representations is Dynamical Movement Primitives (DMPs) which generate reproductions from integrating a dynamical system~\cite{pastorDMP2009}. Solving DMPs can also be seen as an instance of norm minimization~\cite{dragan2015movement}. DMPs reproductions have desirable features such as smoothness and robustness against perturbations. Although DMPs rely on a single demonstration, their extensions, such as Correlated DMPs (CorrDMPs)~\cite{calinon2010learning_CorrDMP} or Probabilistic Movement Primitives (ProMPs)~\cite{Paraschos2013ProMP}, incorporate multiple demonstrations. We compare our approach against DMPs, CorrDMPs, and ProMPs.


Among others, some LfD approaches fit a statistical model to the given set of demonstrations~\cite{Calinon2007GMM}. One of the most popular statistical methods uses a combination of Gaussian Mixture Models (GMMs) and Gaussian Mixture Regression (GMR) and fits the model using the expectation-maximization optimization method~\cite{Calinon2007GMM}. GMM/GMR alone has no ability to satisfy constraints or generalize the model to novel situations. However, GMM/GMR has been extended to incorporate multiple frames which allow for the ability to generalize a reproduction to new conditions~\cite{TPGMMcalinon2016}.

\section{Background on Elastic Maps}
Elastic maps provide a method of optimizing manifolds for several parameters by modeling elastic springs which interconnect nodes and data points. Originally proposed as principal curves and surfaces~\cite{hastie_stuetzle_1989}, elastic maps historically have been used for nonlinear dimensionality reduction~\cite{gorban_zinovyev}. Elastic maps have seen significant use in several applications: predicting election outcomes~\cite{gorban_visualization2001}, constructing principal trees for analyzing gene expressions~\cite{gorban_zinovyev}, and skeletonization of 2-D curves~\cite{piecewiseSkeletonization}. To the best of our knowledge, our proposed approach is the first application of using elastic maps for learning robot skills from demonstrations.

Ideas similar to elastic maps have been proposed in other robotics applications, however. Sumner \emph{et al.,} proposed a shape manipulation algorithm which uses a graph structure and deforms space using this structure~\cite{sumner2007embedded}. The ideas of graph representation and deformation are central to elastic maps. Alternatively, Whelan \emph{et al.,} use energy costs, including a ``spring-like'' energy term, in their optimization for simultaneous localization and mapping (SLAM)~\cite{whelan2016elasticfusion}.


\section{Methodology} 
\label{sec:method}

In this section we explain our proposed method for encoding robot trajectories using elastic maps. The algorithm starts by receiving a set of $m$ demonstrations $\mathcal{X} = \{\vect{X}^1, ..., \vect{X}^m\}$ with $m \geq 1$. An individual demonstration is a discrete finite-length trajectory  $\vect{X}^i = [\vect{x}_1^i, \vect{x}_2^i, ..., \vect{x}_T^i]^{\top} \in \mathbb{R}^{T \times d}$ in robot task-space where $T$ is the number of observed $d$-Dimensional points. We first stack all demonstrations into a time-ordered single vector $\vect{G} = [\vect{x}_1^1, \vect{x}_1^2, ..., \vect{x}_1^m, \vect{x}_2^1, \vect{x}_2^2, ..., \vect{x}_T^m]^{\top} \in \mathbb{R}^{M \times d}$, where $M$ is the total number of data points and $\vect{g}_i$ refers to the $i$-th element of $\vect{G}$. The time-ordered $\vect{G}$ vector allows us to perform the initialization and optimization of the elastic map more efficiently (more details in Sec.~\ref{subsec:optimization}). 


\subsection{Generating a Polyline Elastic Map}
\begin{figure}[h]
\centering
\includegraphics[trim=0 0em 0 0, clip, width=0.8\columnwidth]{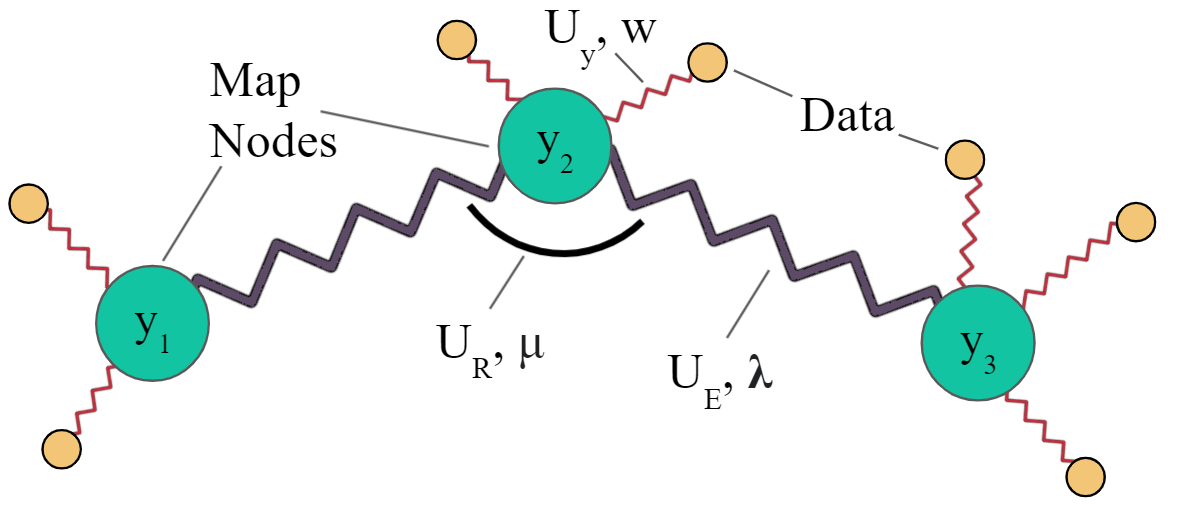}
\caption{\small{Visualization of a simple polyline elastic map and associated energies. This map includes: 3 nodes $y_1$, $y_2$, and $y_3$ with energy $U_Y$; 2 edges $e_1$=\{$y_1$, $y_2\}$ and $e_2$=\{$y_2$, $y_3\}$ with energy $U_E$; and 1 rib $r_1$=\{$y_1$, $y_2$, $y_3$\} with energy $U_R$.}} \label{mapWithEnergy}
\end{figure}

An elastic map is made up of \emph{nodes}, connections between nodes, known as \emph{edges}, and combinations of edges, known as \emph{ribs} (see Fig.~\ref{mapWithEnergy}). Although nodes can have any number of edges, we focus on polyline elastic maps in which each node only has two edges connecting to other nodes, except for terminal nodes, which each have only one edge. An elastic map is first initialized by constructing an array of nodes $\vect{Y} = [\vect{y}_1, \vect{y}_2, ..., \vect{y}_N]^{\top}$, where the number of nodes $N \leq M$ is selected by the user. We examine different methods of constructing the initial maps in Section~\ref{subsec:expinit}. Once these nodes are placed, a set of $P$ edges $\{e_1, e_2, ..., e_P\}$ is found by connecting adjacent pairs of nodes such that $e_i = \{\vect{y}_i, \vect{y}_{i + 1}\}$. Finally, a set of $S$ ribs $\{r_1, r_2, ..., r_S\}$ is found by connecting two adjacent edges as $r_i = \{\vect{y}_{i}, \vect{y}_{i + 1}, \vect{y}_{i + 2}\}$. Note that for polylines, we have $P = N - 1$ and $S = N - 2$.


\subsection{Elastic Map Fitting}
Once the elastic map has been generated the optimal placement of the nodes within the map must be found via minimizing the energy of the map as a whole. There are multiple energy terms that contribute to the overall energy of the map. These energy terms (illustrated in Fig.~\ref{mapWithEnergy}) are
\begin{enumerate*}[label=(\roman*)]
  \item the approximation energy $U_Y$ which measures the fit between nodes and data,
  \item the stretching energy $U_E$ which measures the distance between adjacent nodes, and
  \item the bending energy $U_R$ which measures the curvature of nodes.
\end{enumerate*}
These energy terms can be defined as follows:
\begin{align}
    U_Y &= \frac{1}{\sum_{\vect{g}_j} w_j} \sum_{i = 1}^{N} \sum_{\vect{g}_j \in K_i} w_j || \vect{g}_j - \vect{y}_i ||^2 \label{eq:U_Y}\\
    U_E &= \sum_{i = 1}^{P} \lambda_i || e_i(1) - e_i(0) ||^2  \label{eq:U_E}\\
    U_R &= \sum_{i = 1}^{S} \mu_i || r_i(0) - 2 r_i(1) + r_i(2) ||^2, \label{eq:U_R}
\end{align}
where $K_i$ is the set of data points clustered around a node $\vect{y}_i$, $||\cdot||$ is the $L^2$ norm, $w_j$ is the weight corresponding to the approximation stiffness of data point $\vect{g}_j$, $\lambda_i$ is the weight representing the resistance to stretching of edge $e_i$, and $\mu_i$ is the weight representing the bending resistance of rib $r_i$. The approximation energy $U_Y$ is defined such that it measures the mean squared error between a node and its clustered data points. Minimizing this energy would place the node at the weighted average of the cluster, overall promoting the map to fit demonstrations. The stretching energy $U_E$ is defined such that it measures the Euclidean distance between consecutive nodes. Minimizing this energy results in no distance between adjacent nodes. The bending energy $U_R$ is defined such that it measures the curvature of a node given the adjacent nodes. Minimizing this energy results in no curvature, or all nodes aligned. The process of fitting an elastic map to the observed data can be formulated as the minimization of the total energy as
\begin{equation}
    \underset{\vect{Y}}{\text{minimize }}  \sum_{i \in \{ Y, E, R \}} U_i . \label{eq:opt}
\end{equation}
Assuming $\lambda_i, \mu_i \ge 0$, the objective function in~\eqref{eq:opt} becomes convex, because sum of non-negative convex functions (i.e., $L^2$ norm) is also convex and can be efficiently solved using local optimizers. 
It should be noted that we can simplify the optimization problem even further by distributing the weighting factors $\lambda_i$ and $\mu_i$ uniformly for each edge and rib, resulting in constant weights $\lambda$ and $\mu$. This does not affect results provided no portion of the trajectory requires a different stiffness or bending than any other portion. The constant parameters $\lambda$ and $\mu$ affect the distance between nodes and bending of edges (jerkiness) of the optimized map, respectively. $\lambda$ forces more uniform placement of nodes with more equal edge lengths, but if weighted too highly results in a map where nodes prioritize proximity to each other over data points. $\mu$, on the other hand, incentivizes a smoother map with overall lower curvature and jerk, and if weighted too highly forces nodes into a line, resulting in a loss of curvature of the fit. 

The individual weights $w_i$ of data points strongly affect the resulting map. We examine different methods of determining these weights based on trajectory features in Section~\ref{subsec:expinit}. Additionally, these weights can be used to insert initial, final, or via-point constraints into the elastic map. A constraint can be added via insertion into $\vect{G}$ at the proper index $i$ with an arbitrarily high corresponding weight $w_i$, forcing the map to directly match that constraint.

\begin{figure}[b]
\centering
\includegraphics[trim=0 0em 0 0, clip, width=0.7\columnwidth]{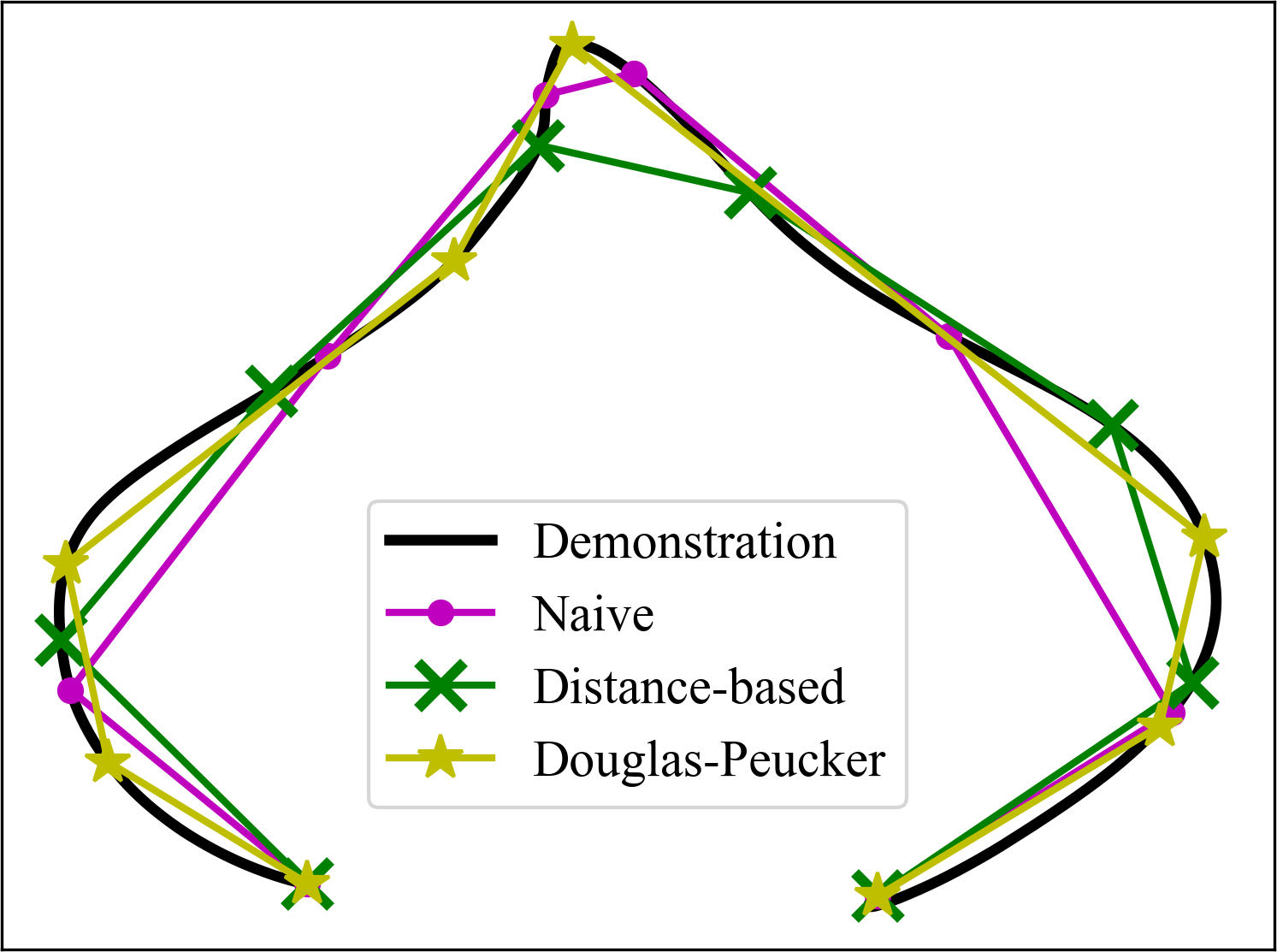}
\caption{\small{Three downsampling methods examined for initialization of elastic maps on a 2D handwriting demonstration.}}
\label{downsample_demo}
\end{figure}

\subsection{Initialization Methods}
\label{subsec:init}
The first step of generating an elastic map is the initial placements of the nodes. In previous works, this is done at random, or evenly distributed amongst the data~\cite{gorban_visualization2001}. For our approach, we leverage prior knowledge of the trajectories to find an initial placement, resulting in quicker, easier, and more accurate optimization. We examine three downsampling methods: naive, distance-based, and Douglas-Peucker~\cite{douglas_peucker1973algorithms}. Naive downsampling uses the first element and every $\frac{M - 1}{N - 1}$ element of $\vect{G}$ afterwards to initialize the nodes $\vect{Y}$. This results in an initial map which is evenly spaced in step size and suitable for minimizing the objective $U_Y$. Distance-based downsampling finds $N$ elements of $\vect{G}$ (including the first and last elements) with distance $\Delta = \frac{L}{N - 1}$ between each element, where $L$ is the total curve length of $\vect{G}$. This method gives an initial map which has even spacing between all nodes and suitable for minimizing the objective $U_E$. The Douglas-Peucker downsampling~\cite{douglas_peucker1973algorithms} algorithm begins with a node at each end of the trajectory and a single connecting edge, then iteratively locates the data point $\vect{g}_i$ for which the distance from $\vect{g}_i$ to the current set of nodes is the maximum, and places the next node at $\vect{g}_i$. This process repeats until no point $\vect{g}_i$ lies outside of a certain distance \bm{$\varepsilon$} of the current trajectory. This method matches high jerk features better than other methods, and is thus well suited for minimizing the objective $U_R$. An example of employing these  methods can be seen in Fig.~\ref{downsample_demo} where different downsampling methods with $M=1000$ and $N=8$ are used on a 2D trajectory from the LASA dataset~\cite{Khansari-Zadeh2011LASA}.

\subsection{Weighting Methods}
\label{subsec:weight}
The next step is weight assignment where a weight value $w_i$ must be assigned to each data point $\vect{g}_i$. Manual weight assignment can be impractical for human users, so in this section we study weighting schemes. More specifically, we examine three weighting schemes: uniform, curvature-based, and jerk-based. In the uniform weighting scheme, all weights are set to the same value such that $w_1 = w_2 = ... = w_M = h$, where $0 < h \leq 1$. The curvature-based weighting scheme assigns weight to data point $\vect{g}_i$ based on the curvature at that point, as $w_i = || \vect{x}_{j - 1}^{k} - 2 \vect{x}_j^k + \vect{x}_{j + 1}^{k} ||$, where $||\cdot||$ is the $L^2$ norm and $\vect{x}_j^k$ is the point corresponding to $\vect{g}_i$ and weight $w_i$. This weighting scheme preserves high curvature features from given trajectories. The jerk-based weighting scheme assigns weight $w_i$ to data point $\vect{g}_i$ based on the jerk at that point, resulting in $w_i= ||-\vect{x}_{j - 2}^{k} + 2 \vect{x}_{j - 1}^{k} - 2 \vect{x}_{j + 1}^{k} + \vect{x}_{j + 2}^{k} ||$. This weighting scheme preserves high jerk features of given trajectories. In Section~\ref{subsec:init_weight_test}, we examine the proposed initialization and weighting schemes using multiple metrics.

\subsection{Optimizing with Expectation-Maximization}
\label{subsec:optimization}
Once the initial map has been generated, we use a local optimizer to solve~\eqref{eq:opt}. Among other optimization methods, in this paper, we chose Expectation-Maximization (EM). We specifically employ a modified EM algorithm developed for elastic maps as presented in \cite{gorban_visualization2001}.


As shown in Algorithm 1, the EM algorithm consists of two main steps; in the expectation step, each data point $\vect{g}$ is placed in cluster $K_i$ corresponding to the closest node $\vect{y}_i$ based on the Euclidean distance. In the maximization step, the placements of the nodes are adjusted such that the energy of the map is minimized according to the current clustering. These steps are repeated until convergence is achieved.


Because our elastic map is an ordered polyline, some steps of the original algorithm can be simplified. Given an initial placement of nodes and the data to be modeled with corresponding weights, we first initialize two ${N \times N}$ matrices $\vect{E}$ and $\vect{S}$ as
\begin{equation}
\vect{E} = \lambda
\left[\begin{smallmatrix}
 1 & 1 &  0 & \cdots & \cdots & 0 \\
1 &  2 & 1 & 0 & \cdots & 0 \\
0 & 1 &  2 & 1 & \cdots  & 0 \\
\vdots & \ddots & \ddots & \ddots & \ddots & \vdots \\
0 & \cdots & 0 & 1 &  2 & 1\\
0 & \cdots &  \cdots & 0 & 1 & 1   \\
\end{smallmatrix}\right], \nonumber
\vect{S} = \frac{\mu}{4}
\left[\begin{smallmatrix}
1 & -2 &  1 & 0 & \cdots &\cdots & 0 \\
-2 & 5 & -4 & 1 & 0 &\cdots & 0 \\
1 & -4 & 6 & -4 & 1 & \cdots & 0 \\
0 & 1 & -4 & 6 & -4 & \cdots & 0 \\
\vdots & \ddots & \ddots & \ddots & \ddots & \ddots & \vdots\\
0 & \cdots & 1 & -4 & 6 & -4 & 1\\
0 & \cdots & 0 &  1 & -4 & 5 & -2\\
0 & \cdots & \cdots & 0 & 1 & -2 & 1   \\
\end{smallmatrix}\right]
\end{equation}
where $\vect{S}$ represents the forces of each rib's bending on the position of each map node, and $\vect{E}$ represents the pulling force of each edge on its neighboring nodes. The elements of these matrices are derived from equations \eqref{eq:U_E} and \eqref{eq:U_R}, where rows/columns are nodes and elements represent connections between nodes. Next, we perform the expectation step by assigning every data point $\vect{g}_j$ to the closest node $\vect{y}_i$ and denoting this cluster $K_i$ (lines 4-6 in Algorithm 1). Then for each $K_i$ we compute the pulling force of each data point on its assigned cluster node, $\vect{V}$, which is a diagonal matrix with element $\vect{V}_{ii} = \sum_{\vect{g}_j\in{K_i}}w_j$. Together, these matrices can be summed to find the matrix $\vect{A}$, as
\begin{equation}
   \vect{A} = \frac{1}{\sum_{\vect{g}_j} w_j}\vect{V} + \vect{E} + \vect{S} \label{eq:aij}
\end{equation}
which represents the total net forces pulling on each node. 
Next we construct the vector $\vect{C} = [c_1, ..., c_N]^\top$ where $c_i=\sum_{\vect{g}_k\in{K_i}}w_k \vect{g}_k / \sum_{k=1}^{M} w_k$, the weighted center of all data points in cluster $K_i$. Finally, we solve the following system of equations $\vect{A} \vect{Y} = \vect{C}$ resulting in a new and more optimal node positions $\vect{Y}$*. Since $\vect{A}$ is square and full rank, the solution $\vect{Y}$* is exact. Once $\vect{Y}$* is found, we repeat the cluster assignment and solve for a new $\vect{Y}$* iteratively until new positions do not differ from old positions, yielding a locally optimal placement of nodes. We have tested other local and global optimization solvers as alternatives to the EM algorithm; our results found that the global solvers are far slower to converge to the same optimum.

\begin{algorithm}[t]
\DontPrintSemicolon
\label{EM_alg}
  \KwInput{$N$, Initial Nodes $\vect{Y}$, Data $\vect{G}$, Weights $\vect{w}$, $\lambda$, $\mu$}
  \KwOutput{Optimal Nodes $\vect{Y}$*}

   Initialize $\vect{E}$, $\vect{S}$, set $b = \sum \vect{w}$\; 
   \While{not converged}
   {
        $K_i$ = [] for $i = 1...N$
        \tcp{Expectation}
        \For{$\vect{g} \in \vect{G}$}
        {
            $j = \argmin_{i=1...N} ||\vect{g}$ - $\vect{y}_i||$\;
            $K_j \longleftarrow$ $\vect{g}$\;
        }
        $\vect{A} = \vect{E} + \vect{S}$ \tcp{Maximization}
        $\vect{C}$ = []\;
        \For{$i=1...N$}
        {
            $\vect{A}_{ii}$ = $\vect{A}_{ii} + (\sum_{\vect{g}_j \in K_i} w_j)/b$\;
            $\vect{C}_i = (\sum_{\vect{g}_j \in K_i} w_j \vect{g}_j)/b$\;
        }
         $\vect{Y}$* = $\vect{A}^{-1}\vect{C}$
        
   }

\caption{Elastic map EM algorithm}
\end{algorithm}


\section{Experiments}
\label{sec:exps}
We evaluate our approach in a variety of experiments and verify the feasibility of learning from demonstration using elastic maps both in simulation and on a real-world robot. We examine how different methods for building the elastic maps affect the reproduced trajectories and the effects of the parameters used to construct the elastic maps based on computation time and properties of the resulting reproduction. Convergence to the demonstration(s), curvature, and smoothness are compared. Additionally, we compare the proposed approach to three other contemporary LfD methods using the same metrics. All experiments are performed using code in Python\footnote{ \url{https://github.com/brenhertel/ElMapTrajectories}} running on a single thread of an AMD Ryzen 5 3600 CPU at 3.6 GHz.

\subsection{Elastic Maps for Trajectory Reproduction}
\label{subsec:elmap_basic}
\begin{figure}[t]
\centering
\includegraphics[trim=0 0em 0 0, clip, width=0.9\columnwidth]{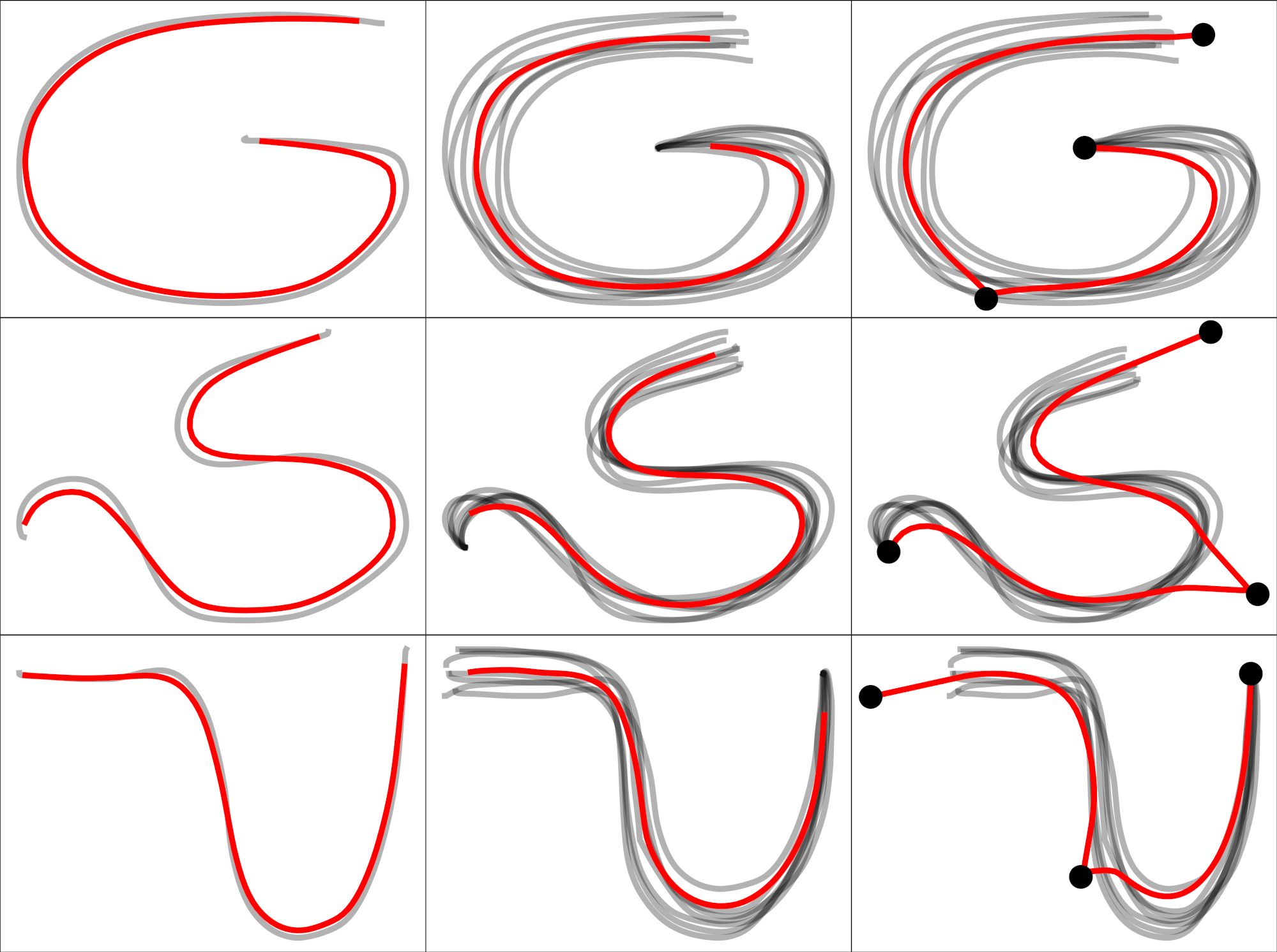}
\caption{\small{Elastic map reproductions ($\color{red} \bm{-} $) with different numbers of demonstrations ($\color{gray} \bm{-} $) and constraints ($\bullet$) for various handwriting shapes.}} \label{elmap_lasa}
\end{figure}

We first verify the effectiveness of elastic maps for modeling 2D handwriting demonstrations from the LASA dataset~\cite{Khansari-Zadeh2011LASA}. As shown in Fig.~\ref{elmap_lasa}, three experiments for each shape were performed using different numbers of demonstrations and constraints. For all experiments, $N = 100$, $\lambda = 0.1$, $\mu = 0.01$, and curvature-based weighting as well as distance-based initialization were used. The first column of Fig.~\ref{elmap_lasa} shows how elastic maps (red) perform when a single demonstration (gray) was given. It can be seen that single demonstrations are sufficient for providing an accurate reproduction using our LfD approach. In the middle column, we show the elastic map reproductions found from multiple demonstrations. This shows the flexibility of elastic maps, which can be used with one or more demonstrations. Finally, the third column uses multiple demonstrations and initial, final, and via-point constraints (black dots). The reproduction adheres to these constraints while accurately modeling the provided trajectories. This shows how our LfD approach incorporates given constraints.

\subsection{Effect of the Number of Nodes: Time vs. Dissimilarity}
\label{subsec:expinit}
\begin{figure}[h]
\centering
\includegraphics[trim=0 0em 0 0, clip, width=0.75\columnwidth]{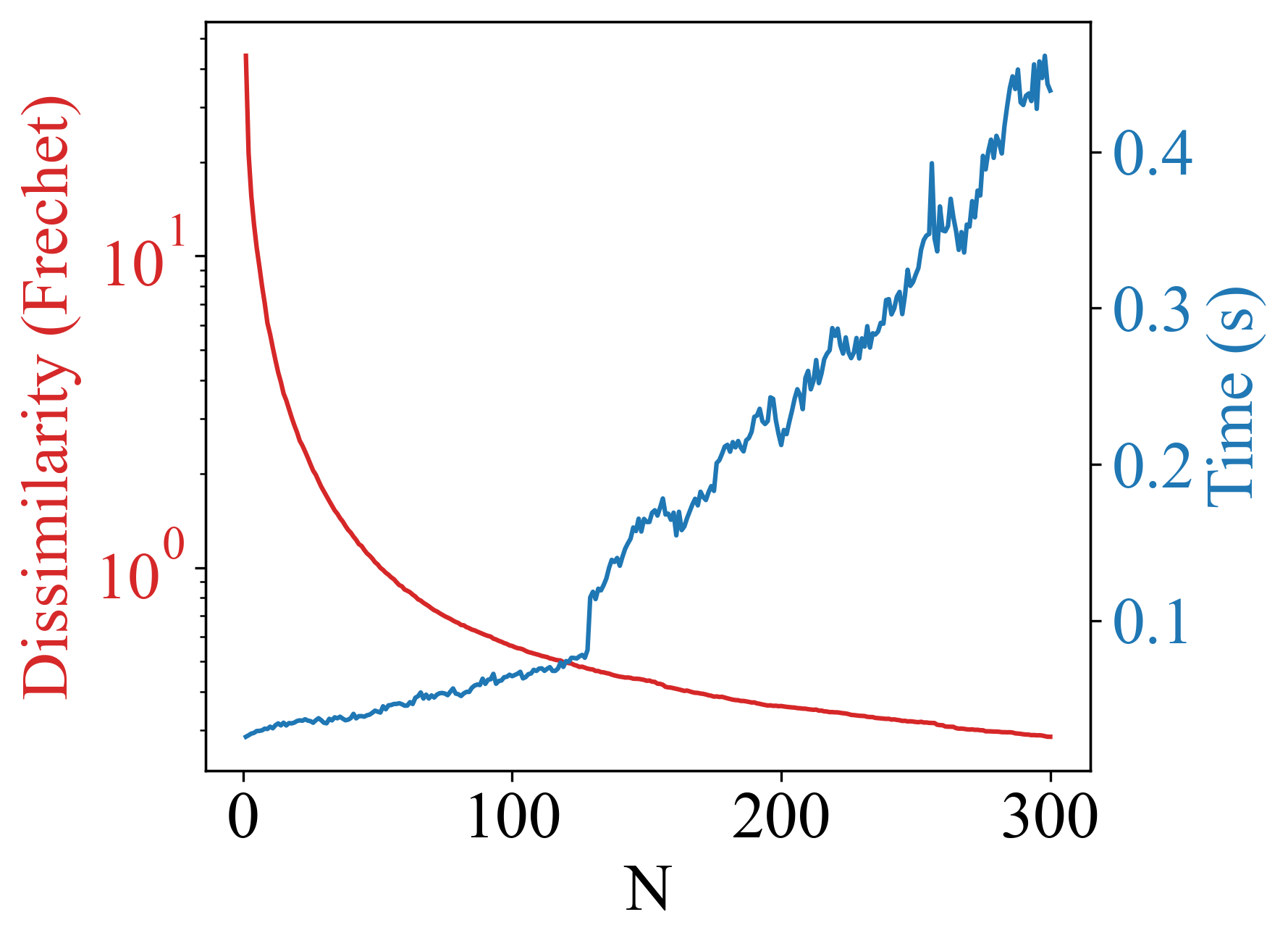}
\caption{\small{Effect of the number of nodes $N$ on computation time ($\color{bblue} \bm{-} $) and dissimilarity ($\color{red} \bm{-} $). Note the log scale on the dissimilarity axis.}} \label{N_experiment}
\end{figure}

The number of nodes needed to represent the elastic map can vary upon the application. For trajectory generation, we observe how varying $N$ can affect the time needed to generate the elastic map and the similarity (or dissimilarity) of the reproduced trajectory. We vary $N$ from 1 to 300, generating a reproduction for each $N$ value for one demonstration from each of the 26 shapes in the LASA dataset~\cite{Khansari-Zadeh2011LASA} (with $M = 1000$). We  record the time taken to generate the reproductions and the dissimilarity between each reproduction and its corresponding demonstration, measured by the discrete Fr\'echet distance \cite{Eiter1994frechet_computing}. For this experiment, we set $\lambda = 0.01$, $\mu = 0.001$, and the elastic map was built with distance-based initialization and curvature-based weighting. The results can be seen in Fig.~\ref{N_experiment}, which depicts computation time and dissimilarity averaged over the 26 separate experiments. As expected, computation time increases with $N$ while dissimilarity decreases. However, the rate at which these two properties change varies widely. From this point onward, we set $N = 100$ because it maintains low computation time and dissimilarity.

\subsection{Effects of Initialization and Weighting Methods}
\label{subsec:init_weight_test}
Two of the most configurable aspects of LfD with elastic maps are the initial placement of the nodes and the weighting of each data point $w_i$. We proposed different methods of each aspect in Sections~\ref{subsec:init} and \ref{subsec:weight}. In this experiment, we study each combination against several similarity measures to determine which method performs best for generating reproductions. For this experiment, we used a single demonstration from each of the 26 shapes in~\cite{Khansari-Zadeh2011LASA}. We measured the resulting computation time, Fr\'echet distance, angular dissimilarity~\cite{ontanon2020overview}, and the total jerk value of the reproduced trajectories. In these experiments, we set $N = 100$, $\lambda = 0.01$ and $\mu = 0.001$. The results are reported in Table~\ref{tab:all-table}. Although no single combination of methods definitively outperforms the others, the combination of distance-based downsampling and curvature-based weighting provides good results and is used throughout all other experiments. 


\begin{table}[h]
\setlength{\tabcolsep}{3.5pt}
    \centering
    \caption{\small{Results of examining different combinations of initialization and weighting methods, measured across the computation time (seconds), Fr\'echet distance in Cartesian and curvature spaces, and total jerk of the reproduction, averaged over 26 experiments.}}
    \begin{tabular}{l|ccc|ccc}
    \toprule
    & \multicolumn{3}{c}{Time (s)} & \multicolumn{3}{|c}{Fr\'echet}\\ \midrule
    & Naive & \multicolumn{1}{c}{\begin{tabular}[c]{@{}c@{}}Distance- \\ based\end{tabular}} & \multicolumn{1}{c|}{\begin{tabular}[c]{@{}c@{}}Douglas- \\ Peucker\end{tabular}} & Naive & \multicolumn{1}{c}{\begin{tabular}[c]{@{}c@{}}Distance- \\ based\end{tabular}} & \multicolumn{1}{c}{\begin{tabular}[c]{@{}c@{}}Douglas- \\ Peucker\end{tabular}} \\ \midrule
    Uniform   & 0.26 & \textbf{0.06} & 2.13    & 0.28 & 0.26 & 0.32 \\
    Curvature & 0.31 & \textbf{0.06} & 2.10    & 0.23 & \textbf{0.21} & 0.26 \\
    Jerk      & 0.32 & 0.10          & 2.12    & 0.42 & 0.45 & 0.57 \\ \toprule
    & \multicolumn{3}{c}{Angular} & \multicolumn{3}{|c}{Total Jerk}\\ \midrule
    & Naive & \multicolumn{1}{c}{\begin{tabular}[c]{@{}c@{}}Distance- \\ based\end{tabular}} & \multicolumn{1}{c|}{\begin{tabular}[c]{@{}c@{}}Douglas- \\ Peucker\end{tabular}} & Naive & \multicolumn{1}{c}{\begin{tabular}[c]{@{}c@{}}Distance- \\ based\end{tabular}} & \multicolumn{1}{c}{\begin{tabular}[c]{@{}c@{}}Douglas- \\ Peucker\end{tabular}} \\ \midrule
    Uniform   & 0.22 & 0.21 & 0.27             & 0.66 & 0.65 & 0.65 \\
    Curvature & 0.21 & \textbf{0.20} & 0.27    & 0.66 & 0.66 & 0.66 \\
    Jerk      & 0.40 & 0.42 & 0.56             & 0.65 & \textbf{0.64} & \textbf{0.64} \\ \bottomrule
    \end{tabular}
    \label{tab:all-table}
    
\end{table}

\subsection{Real-World Experiments}
\label{subsec:realworld}
\begin{figure}[t]
\centering
\includegraphics[trim=0 0em 0 0, clip, width=0.9\columnwidth]{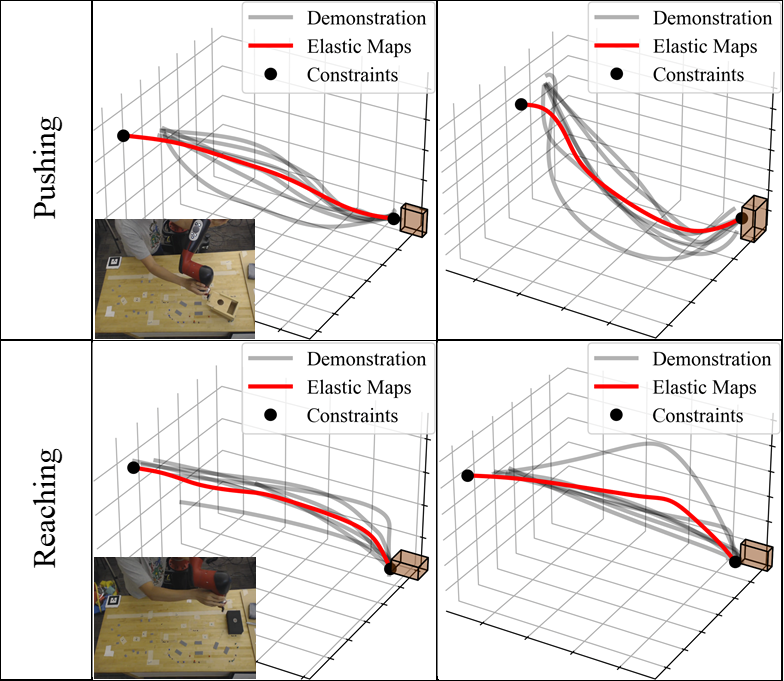}
\caption{\small{Reproductions of two skills in four real-world scenarios.}} \label{rail_demo}
\end{figure}

First, we evaluate the proposed elastic map LfD approach in real-world scenarios using 3D demonstrations of pushing and reaching skills provided by the RAIL dataset~\cite{Rana2020GT_dataset}. We use two scenarios from each skill and multiple demonstrations in each scenario, shown in Fig.~\ref{rail_demo}. We evaluate generalization capability of our approach over initial points by imposing novel initial constraints and similar endpoint constraints. For this experiment, all reproductions were found using $\lambda = 0.01$ and $\mu = 0.1$. Results show that  elastic maps are able to accurately model the given demonstrations and reproduce the real-world skills successfully. Motions of high importance, such as the end of a pushing skill, are modeled successfully and preserved correctly using the proposed approach.

In the next real-world experiment, we model and reproduce a pressing skill using our approach. As depicted in Fig.~\ref{fig1_trajs}, a single demonstration of the task was captured in task space using kinesthetic teaching. We model the captured demonstration using the proposed approach. The reproduced trajectory maintains the key features of the demonstration including the curvature and the smoothness and satisfies the given via-point constraints. For this experiment, we set $\lambda = 0.01$ and $\mu = 0.01$. The generated reproduction is executed on a UR5e 6-DOF manipulator arm using a low-level controller and spherical linear interpolation (Slerp)~\cite{Shoemake1985slerp} for computing orientations. We do not model orientations using elastic maps, and leave this to future work. This reproduction can be seen in the accompanying video.\footnote{Accompanying video at \url{https://youtu.be/rZgN9Pkw0tg}}


\begin{figure}[h]
\centering
\includegraphics[trim=0 0em 0 0, clip, width=0.47\columnwidth]{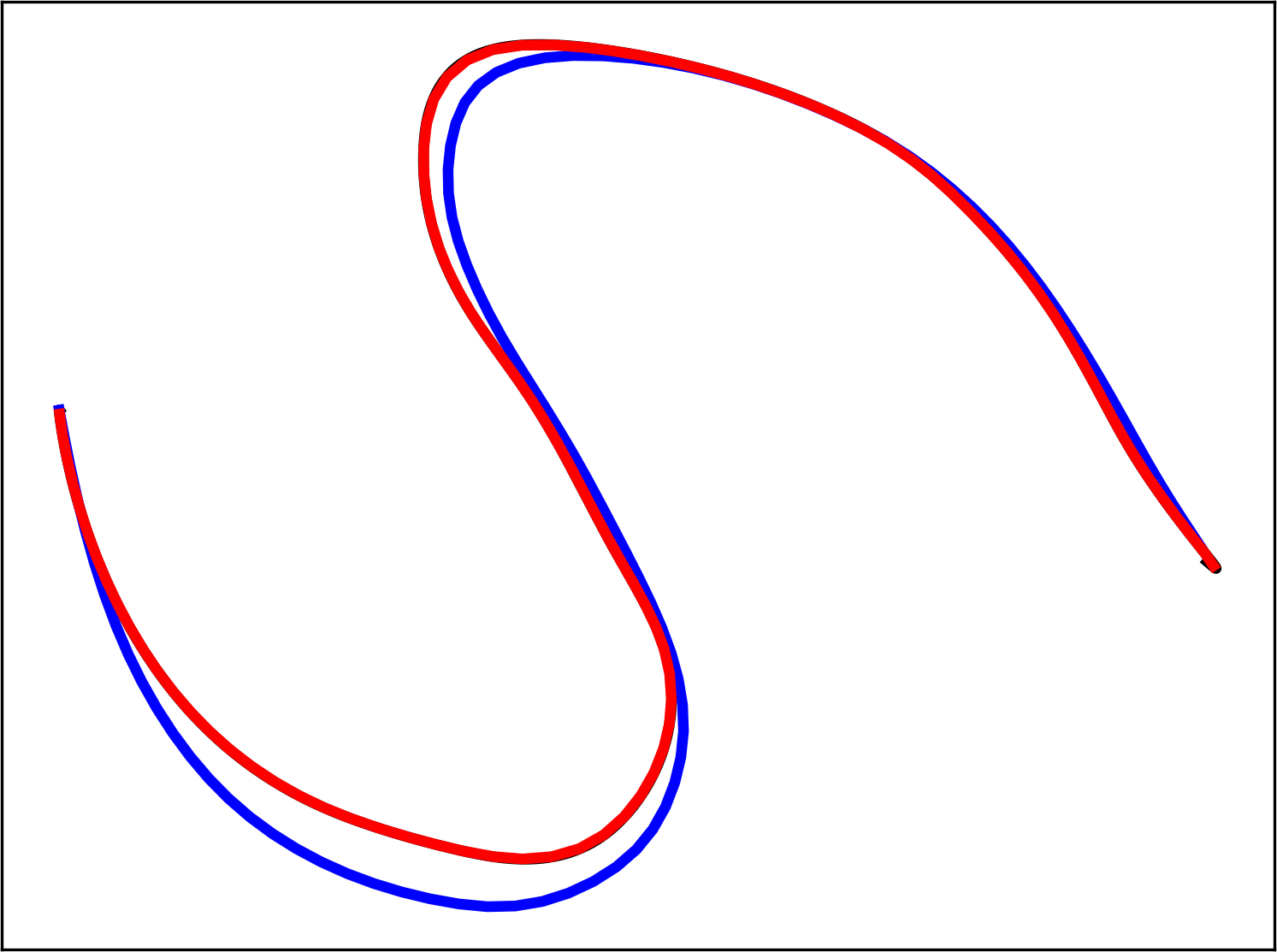}
\includegraphics[trim=0 0em 0 0, clip, width=0.47\columnwidth]{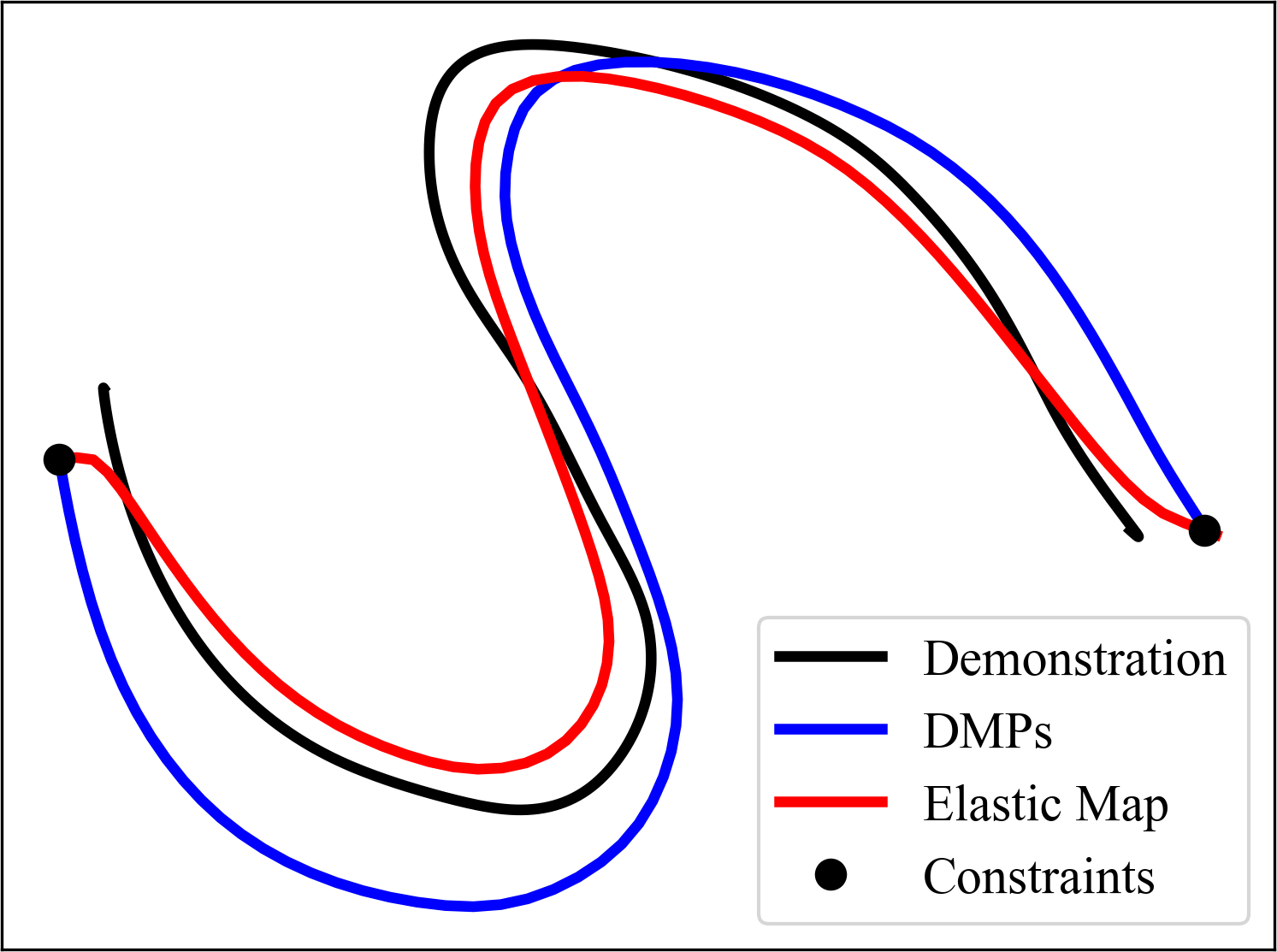}
\caption{\small{Our approach compared to DMPs (tuned to preserve shape) in a 2D writing skill: (right) with, (left) without constraints.}} \label{dmp_cmp}
\end{figure}

\subsection{Comparison to Other LfD Methods}
\label{subsec:cmp}
\begin{figure}[h]
\centering
\includegraphics[trim=0 0em 0 0, clip, width=0.32\columnwidth]{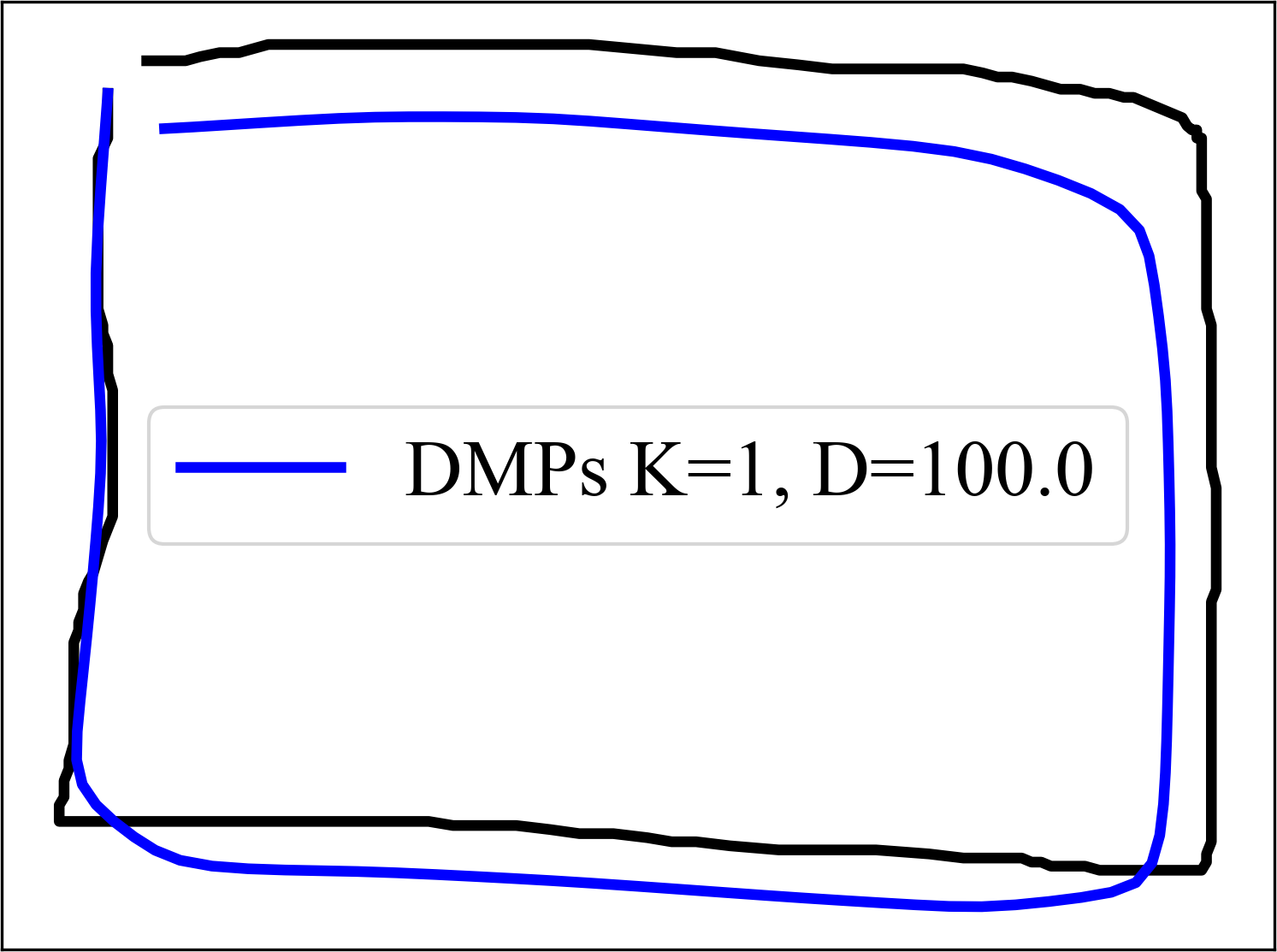}
\includegraphics[trim=0 0em 0 0, clip, width=0.32\columnwidth]{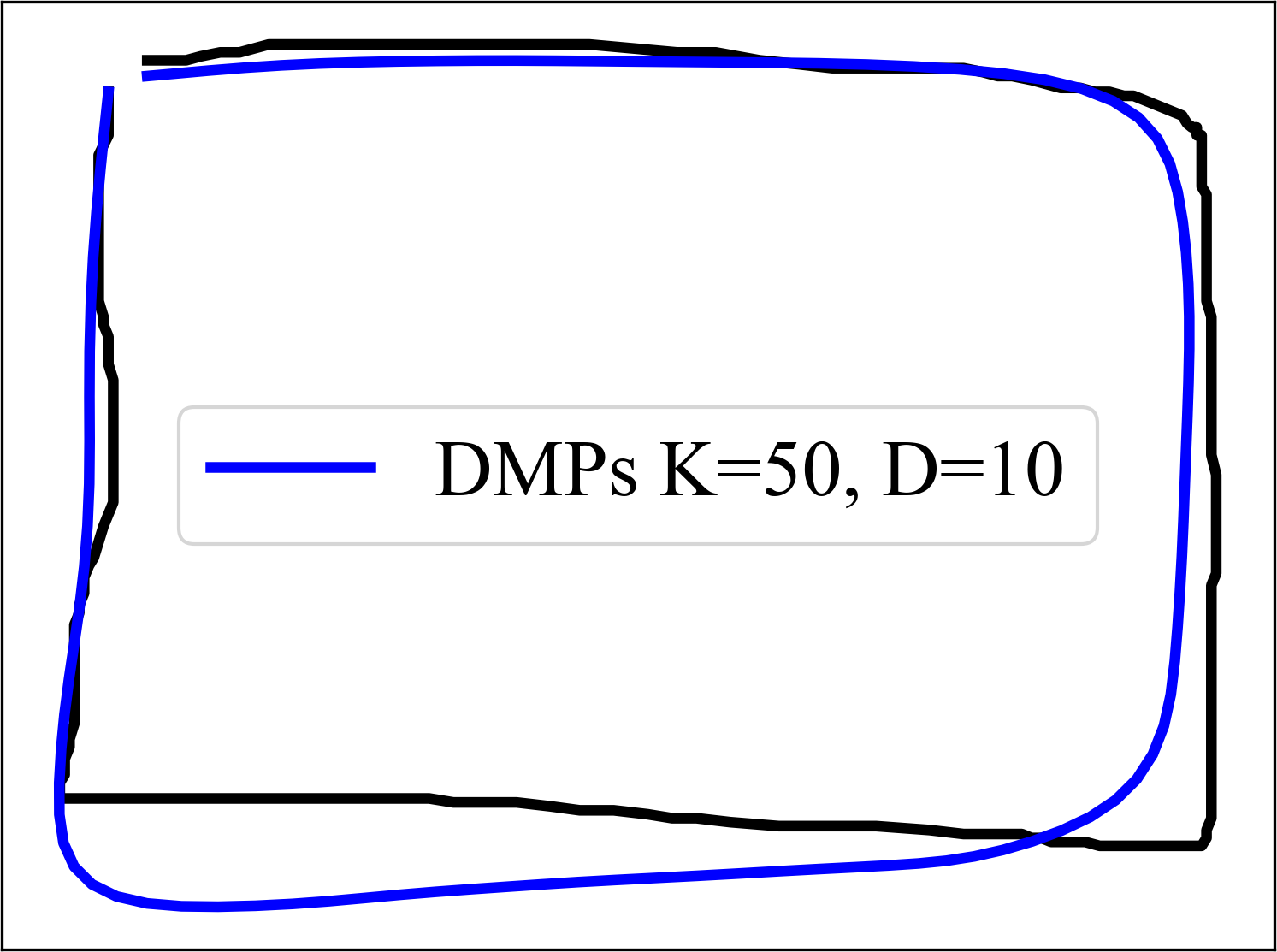}
\includegraphics[trim=0 0em 0 0, clip, width=0.32\columnwidth]{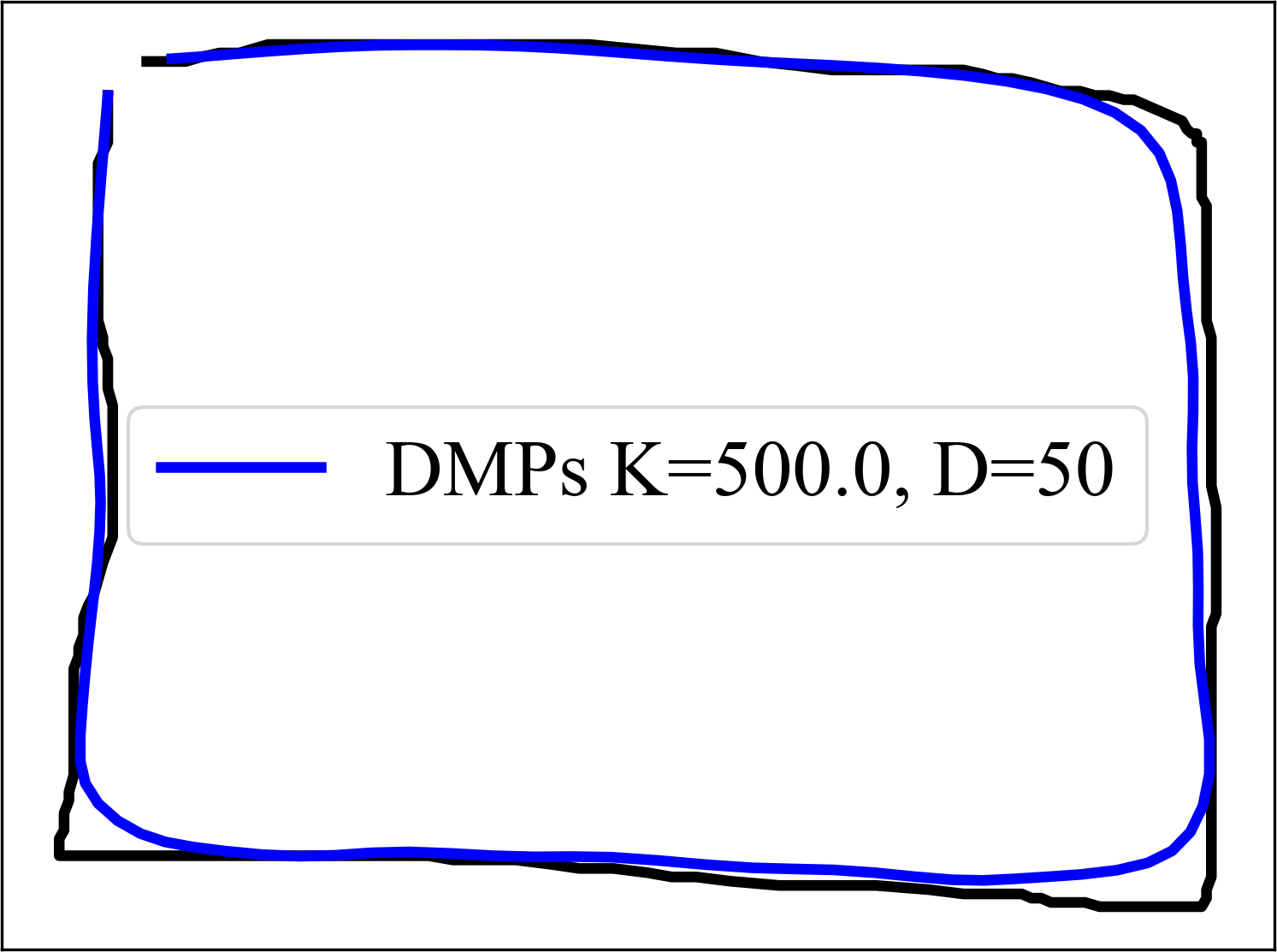}
\includegraphics[trim=0 0em 0 0, clip, width=0.32\columnwidth]{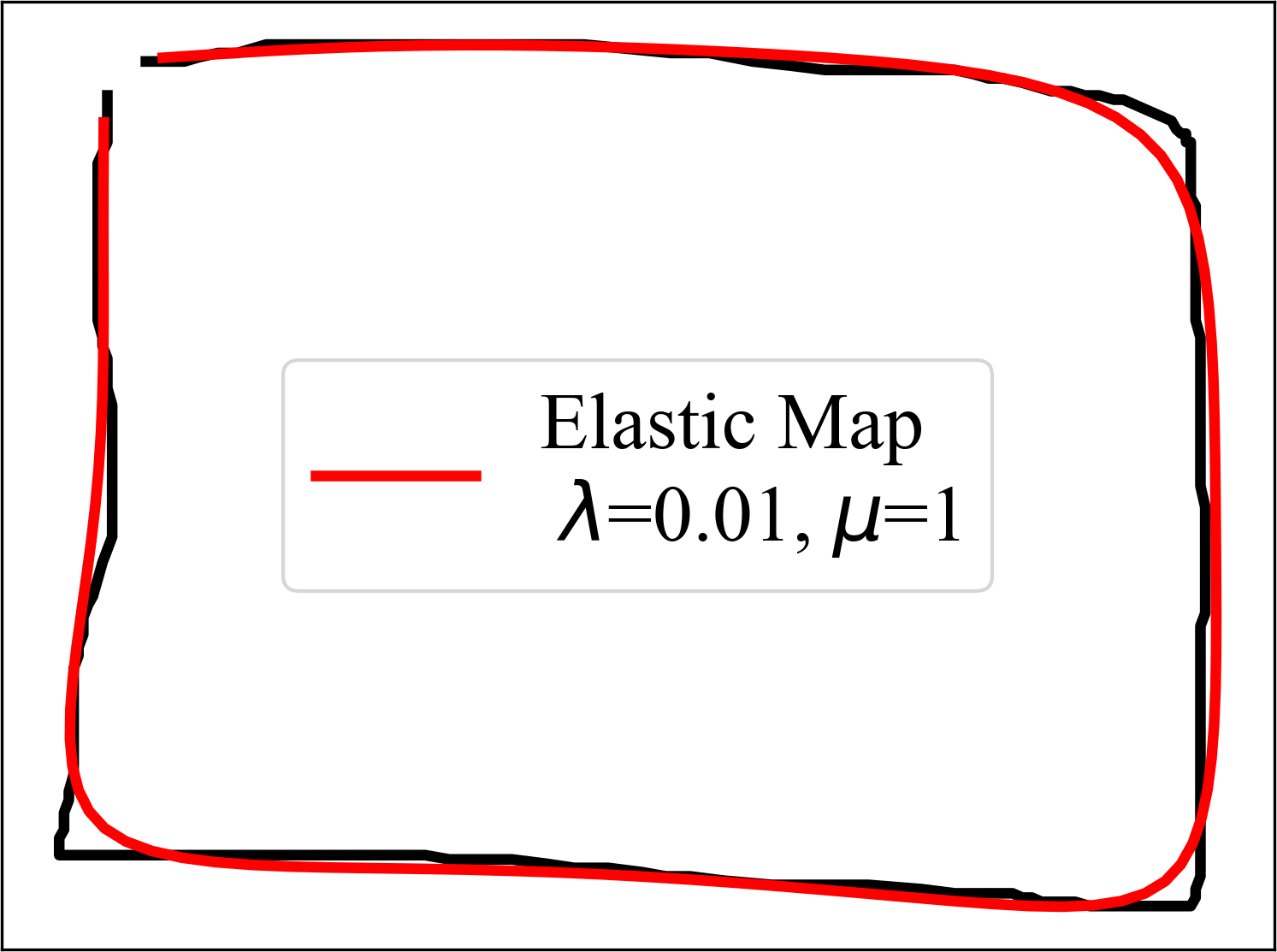}
\includegraphics[trim=0 0em 0 0, clip, width=0.32\columnwidth]{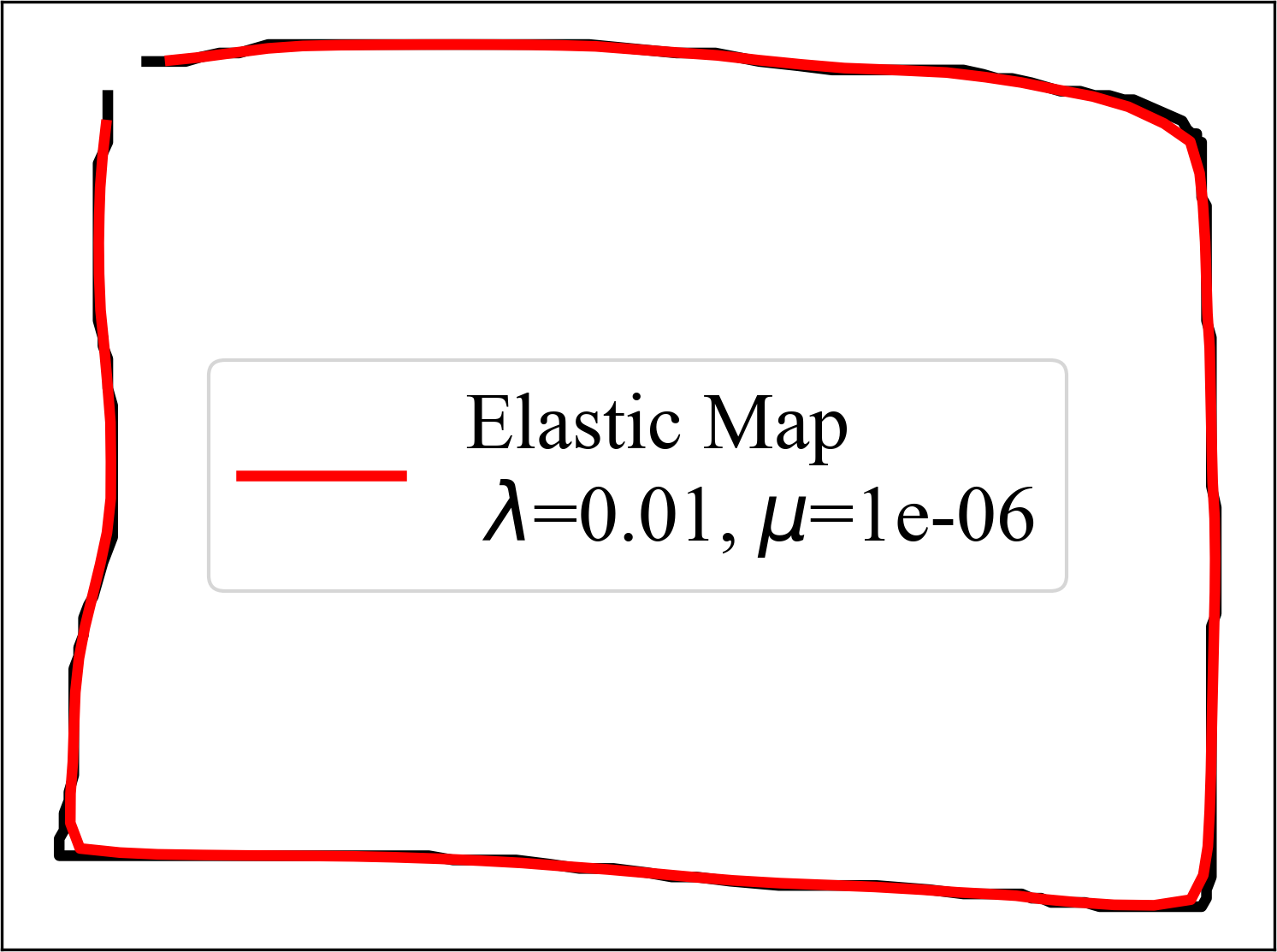}
\includegraphics[trim=0 0em 0 0, clip, width=0.32\columnwidth]{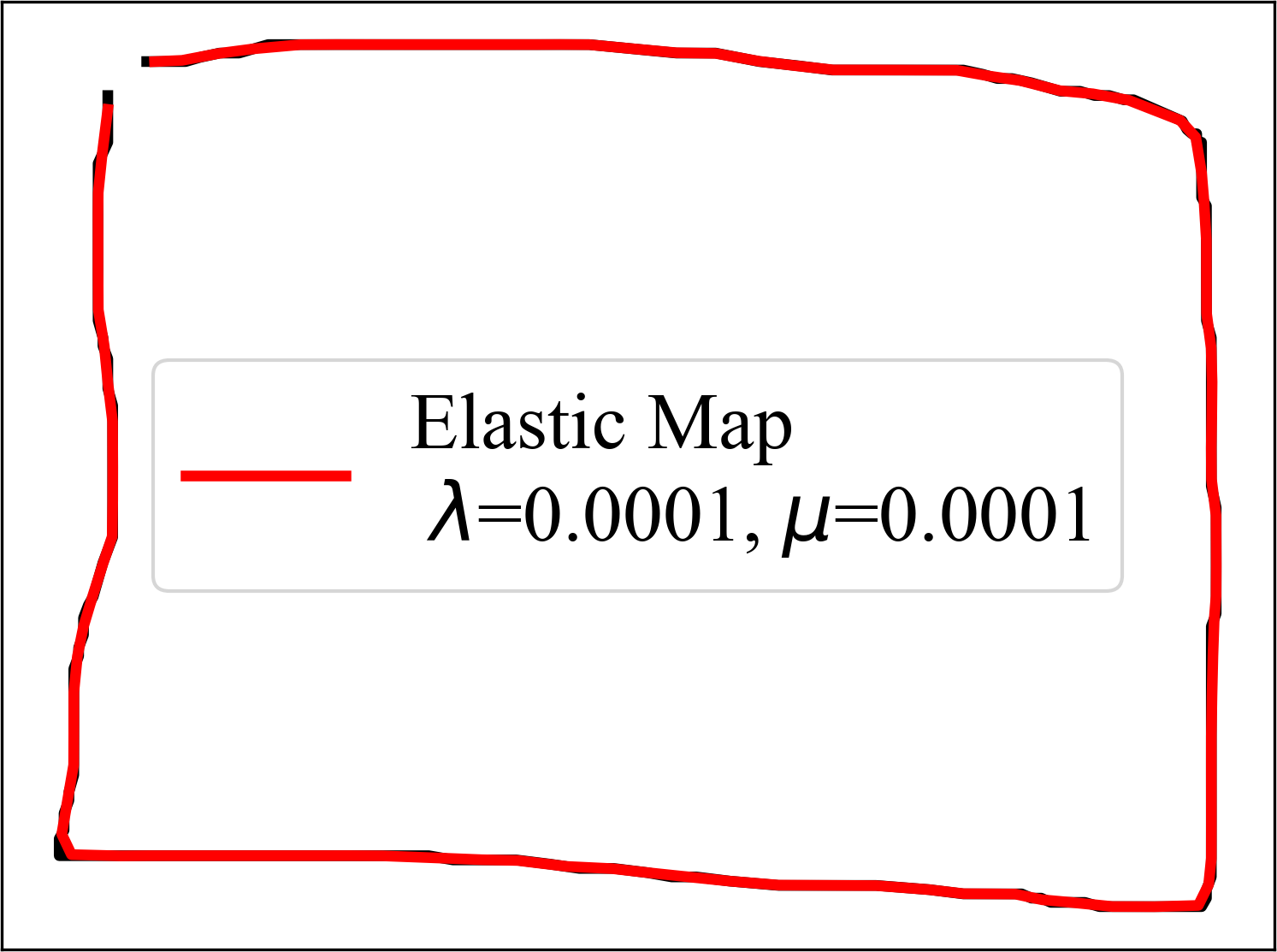}
\caption{\small{Comparison of different tunings for DMPs (top) and elastic maps (bottom). All tunings of DMPs round corners while elastic maps can adhere or round corners depending upon the tuning.}} \label{elmap_dmp_compare_params}
\end{figure}

We compare our approach against four existing LfD approaches. First, we compare using a single demonstration against Dynamic Movement Primitives (DMPs) as formulated in \cite{pastorDMP2009} with 20 basis functions. In this experiment, a 2D writing skill was performed by a human using a pointing device on a screen. The captured demonstration is reproduced by our approach and DMPs in two separate environments, once with constraints and once without, both shown in Fig.~\ref{dmp_cmp}. We measured the performance of the reproductions using the same similarity metrics as Sec.~\ref{subsec:init_weight_test}, with results reported in Table~\ref{tab:dmp-cmp-table}. These results show reproductions using our approach are comparable to DMPs reproductions, and using some metrics, elastic maps even outperform DMPs. Additionally, we perform a qualitative comparison between elastic maps and DMPs shown in Fig.~\ref{elmap_dmp_compare_params}. Here, we show a variety of combination of parameters for both DMPs and elastic maps for a square-like drawing task. We examine different DMPs parameters for the spring constant $K$ and damping coefficient $D$, as well as elastic map parameters for stretching energy $\lambda$ and bending energy $\mu$. It can be seen here that even with highly tuned parameters, DMPs cannot properly model the corners of the box, instead rounding out these features. Elastic maps, on the other hand, can be flexibly tuned to either round or maintain the edges depending on the parameters. This is an important feature in tasks such as robotic welding that requires precision. 

\begin{table}[t]
    \centering
    \caption{\small{Results of comparison on a pushing task between our approach and DMPs in constrained and unconstrained scenarios.}}
    \begin{tabular}{l|cc|cc}
    \toprule
    & \multicolumn{2}{c}{Unconstrained} & \multicolumn{2}{|c}{Constrained}\\ \midrule
    \textbf{Metric}      & Elastic Maps & DMPs & Elastic Maps & DMPs \\ \midrule
    Fr\'echet    & \textbf{0.15}         & 0.46    & \textbf{0.72} & 1.00\\
    Angular      & \textbf{0.03}         & 0.20    & 1.00          & \textbf{0.56}\\
    Jerk         & \textbf{0.95}         & 0.97    & \textbf{0.88} & 1.00\\
    \bottomrule
    \end{tabular}
    \label{tab:dmp-cmp-table}
\end{table}

\begin{figure*}[t]
\centering
\includegraphics[trim=0 0em 0 0, clip, width=0.47\columnwidth]{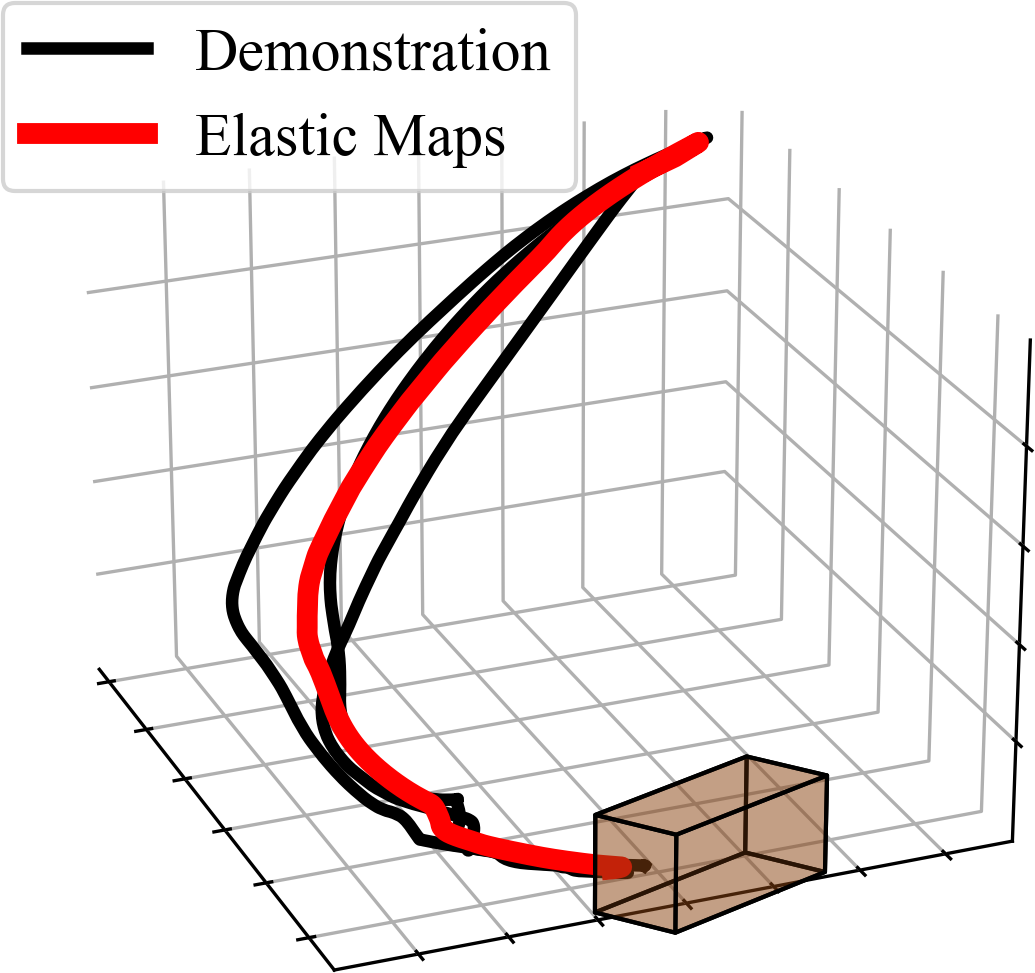}
\includegraphics[trim=0 0em 0 0, clip, width=0.47\columnwidth]{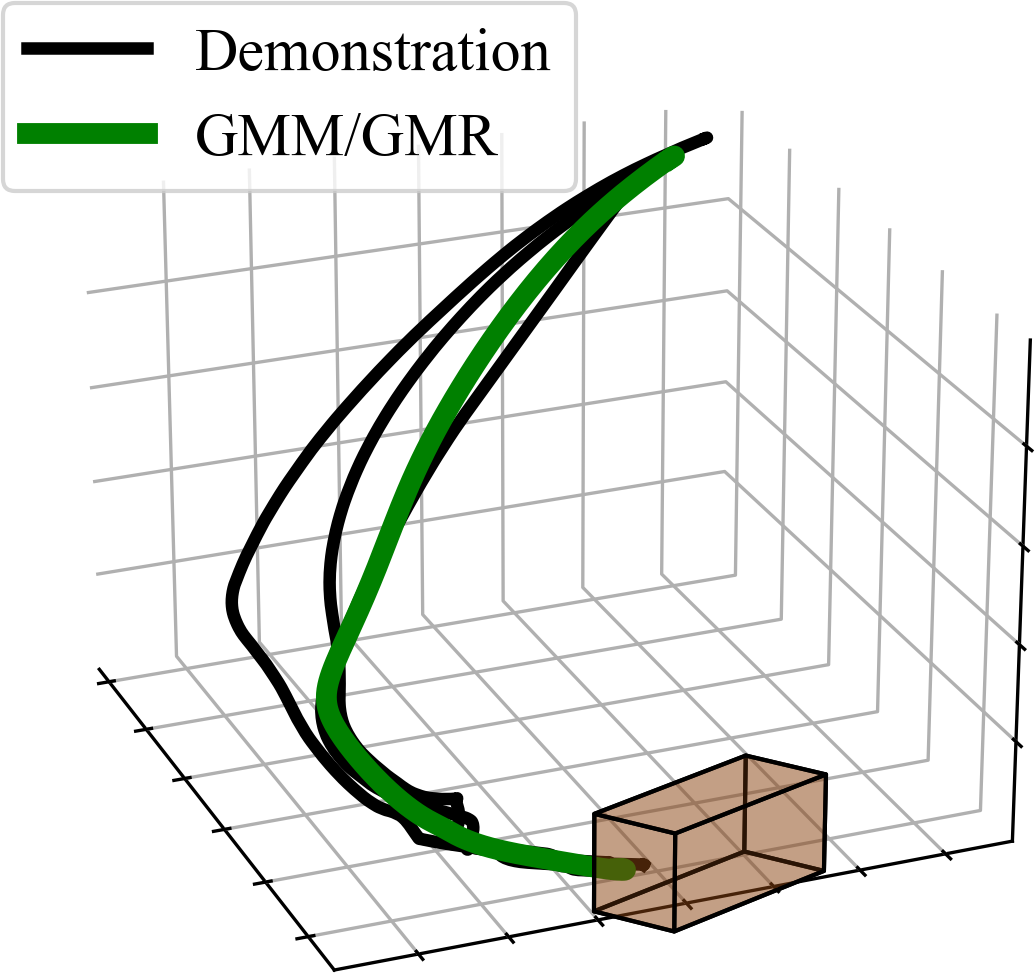}
\includegraphics[trim=0 0em 0 0, clip, width=0.47\columnwidth]{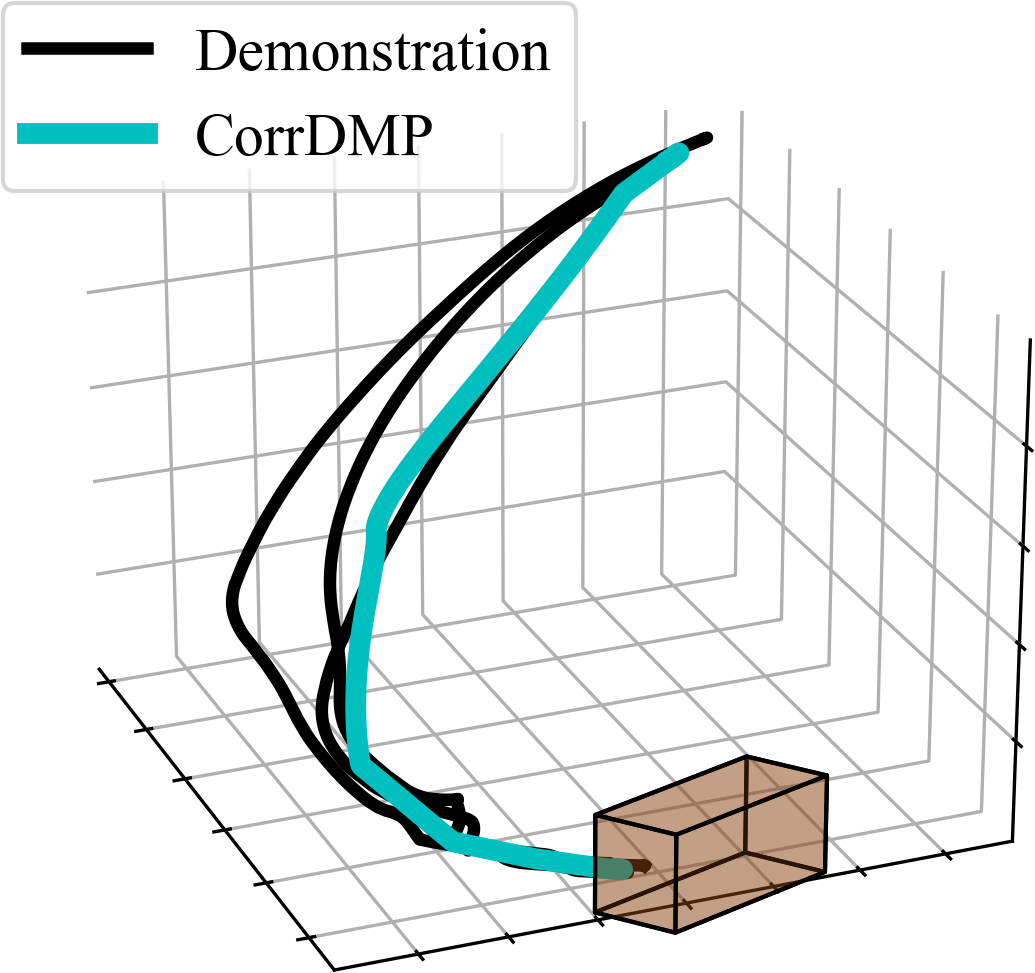}
\includegraphics[trim=0 0em 0 0, clip, width=0.47\columnwidth]{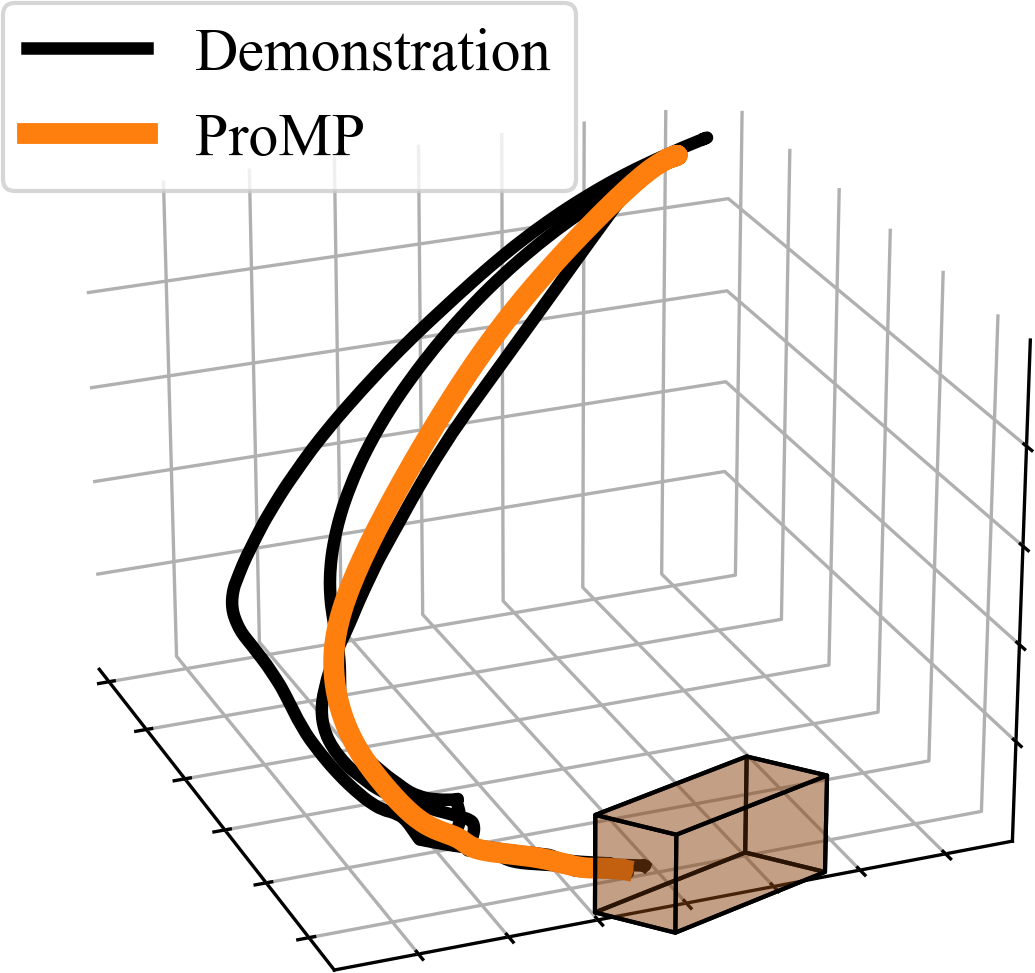}
\caption{\small{Comparison of four LfD approaches for learning and reproduction of a pushing skill.}} \label{corrdmp_gmm_cmp}
\end{figure*}

Next, we compare against three methods which use multiple demonstrations, namely GMM/GMR~\cite{Calinon2007GMM}, Correlated Dynamic Movement Primitives (CorrDMP)~\cite{calinon2010learning_CorrDMP} and Probabilistic Movement Primitives (ProMP)~\cite{Paraschos2013ProMP}. Three demonstrations of a pushing task were captured via kinesthetic teaching on a UR5e manipulator arm. The demonstrations and reproductions from each method can be seen in Fig.~\ref{corrdmp_gmm_cmp}. The results are compared using the same metrics in Sec.~\ref{subsec:init_weight_test}, with results reported in Table~\ref{tab:cmp-table}. All reproductions successfully complete the task, and vary performance among objectives according to the metrics evaluated. These results indicate that using elastic maps for skill learning and reproduction is comparable to other contemporary methods. Additionally, our approach presents other advantages over these methods. It can incorporate any number of initial, final, or via-point constraints which GMM/GMR cannot, unlike CorrDMP does not rely on velocity or acceleration data. Also, it is shown to be well-suited for modeling single or multiple demonstrations.

\begin{table}[h]
    \centering
    \caption{\small{Results of comparison on a pushing task between the proposed approach, GMM/GMR, CorrDMP, and ProMP.}}
    \begin{tabular}{l|cccc}
    \toprule
    \textbf{Metric}      & Elastic Maps & GMM/GMR & CorrDMP  & ProMP\\ \midrule
    Fr\'echet     & 0.85         & \textbf{0.47}    & \textbf{0.47} & \textbf{0.47} \\
    Angular & \textbf{0.60}        & 0.94    & 0.86 & 0.92\\
    Jerk        & 1.00           & 0.87    & \textbf{0.85} & 0.86\\
    \bottomrule
    \end{tabular}
    \label{tab:cmp-table}
\end{table}

\section{Conclusions and Future Work}
\label{sec:conclusion}
We have proposed the use of elastic maps, an existing method for statistical analysis, in the novel application of robotic manipulation via learning from demonstration. Through several experiments, we have showcased the ability for elastic maps to generate reproductions of demonstrated trajectories, and examined methods for determining parameters which are well-suited to the robotics application. Additionally, we compared reproductions generated using elastic maps to those of other contemporary algorithms, and showed that our method has comparable or better performance. Finally, we demonstrated the usability of the proposed approach in real-world pressing task.

Elastic maps present many opportunities to be combined with other LfD representations. Similar to other statistical representations, elastic maps could be used for finding a representative mean of trajectories which can be utilized by some other algorithms. The flexibility and configurability of elastic maps as well as their speed and ability to learn from a single demonstration allow them to be easily applied to various scenarios and show significant potential to be applied in modeling various robot skills for fast and accurate reproductions.

\section*{Acknowledgements}
This work was supported by the U.S. Office of Naval Research (N00014-21-1-2582).

\typeout{}
\bibliographystyle{IEEEtran}
\bibliography{references}

\begin{thebibliography}{10}
\providecommand{\url}[1]{#1}
\csname url@samestyle\endcsname
\providecommand{\newblock}{\relax}
\providecommand{\bibinfo}[2]{#2}
\providecommand{\BIBentrySTDinterwordspacing}{\spaceskip=0pt\relax}
\providecommand{\BIBentryALTinterwordstretchfactor}{4}
\providecommand{\BIBentryALTinterwordspacing}{\spaceskip=\fontdimen2\font plus
\BIBentryALTinterwordstretchfactor\fontdimen3\font minus
  \fontdimen4\font\relax}
\providecommand{\BIBforeignlanguage}[2]{{%
\expandafter\ifx\csname l@#1\endcsname\relax
\typeout{** WARNING: IEEEtran.bst: No hyphenation pattern has been}%
\typeout{** loaded for the language `#1'. Using the pattern for}%
\typeout{** the default language instead.}%
\else
\language=\csname l@#1\endcsname
\fi
#2}}
\providecommand{\BIBdecl}{\relax}
\BIBdecl

\bibitem{pastorDMP2009}
P.~Pastor, H.~Hoffmann, T.~Asfour, and S.~Schaal, ``Learning and generalization
  of motor skills by learning from demonstration,'' in \emph{2009 IEEE
  International Conference on Robotics and Automation}.\hskip 1em plus 0.5em
  minus 0.4em\relax IEEE, 2009, pp. 763--768.

\bibitem{Calinon2007GMM}
S.~Calinon, F.~Guenter, and A.~Billard, ``On learning, representing, and
  generalizing a task in a humanoid robot,'' \emph{IEEE Transactions on
  Systems, Man, and Cybernetics, Part B (Cybernetics)}, vol.~37, no.~2, pp.
  286--298, 2007.

\bibitem{Paraschos2013ProMP}
A.~Paraschos, C.~Daniel, J.~R. Peters, and G.~Neumann, ``Probabilistic movement
  primitives,'' in \emph{Advances in neural information processing systems},
  2013, pp. 2616--2624.

\bibitem{Ahmadzadeh2018TLGC}
S.~R. Ahmadzadeh and S.~Chernova, ``Trajectory-based skill learning using
  generalized cylinders,'' \emph{Frontiers in Robotics and AI}, vol.~5, p. 132,
  2018.

\bibitem{hertel2021SAMLfD}
B.~Hertel and S.~R. Ahmadzadeh, ``Similarity-aware skill reproduction based on
  multi-representational learning from demonstrations,'' in \emph{20th
  International Conference on Advanced Robotics (ICAR)}.\hskip 1em plus 0.5em
  minus 0.4em\relax IEEE, 2021.

\bibitem{calinon2010learning_CorrDMP}
S.~Calinon, I.~Sardellitti, and D.~G. Caldwell, ``Learning-based control
  strategy for safe human-robot interaction exploiting task and robot
  redundancies,'' in \emph{2010 IEEE/RSJ International Conference on
  Intelligent Robots and Systems}.\hskip 1em plus 0.5em minus 0.4em\relax IEEE,
  2010, pp. 249--254.

\bibitem{gorban_zinovyev}
A.~N. Gorban and A.~Y. Zinovyev, ``Principal graphs and manifolds,''
  \emph{Handbook of Research on Machine Learning Applications and Trends}, p.
  28–59, 2008.

\bibitem{Meirovitch2016JA}
Y.~Meirovitch, D.~Bennequin, and T.~Flash, ``Geometrical invariance and
  smoothness maximization for task-space movement generation,'' \emph{IEEE
  Transactions on Robotics}, vol.~32, no.~4, pp. 837--853, 2016.

\bibitem{zucker2013chomp}
M.~Zucker, N.~Ratliff, A.~D. Dragan, M.~Pivtoraiko, M.~Klingensmith, C.~M.
  Dellin, J.~A. Bagnell, and S.~S. Srinivasa, ``Chomp: Covariant hamiltonian
  optimization for motion planning,'' \emph{The International Journal of
  Robotics Research}, vol.~32, no. 9-10, pp. 1164--1193, 2013.

\bibitem{Ravichandar2019MCCB}
H.~Ravichandar, S.~R. Ahmadzadeh, M.~A. Rana, and S.~Chernova, ``Skill
  acquisition via automated multi-coordinate cost balancing,'' in \emph{2019
  International Conference on Robotics and Automation (ICRA)}.\hskip 1em plus
  0.5em minus 0.4em\relax IEEE, 2019, pp. 7776--7782.

\bibitem{hertel2021TLFSD}
B.~Hertel and S.~R. Ahmadzadeh, ``Learning from successful and failed
  demonstrations via optimization,'' in \emph{IEEE/RSJ International Conference
  on Intelligent Robots and Systems (IROS)}.\hskip 1em plus 0.5em minus
  0.4em\relax IEEE, 2021.

\bibitem{dragan2015movement}
A.~D. Dragan, K.~Muelling, J.~A. Bagnell, and S.~S. Srinivasa, ``Movement
  primitives via optimization,'' in \emph{2015 IEEE International Conference on
  Robotics and Automation (ICRA)}.\hskip 1em plus 0.5em minus 0.4em\relax IEEE,
  2015, pp. 2339--2346.

\bibitem{TPGMMcalinon2016}
S.~Calinon, ``A tutorial on task-parameterized movement learning and
  retrieval,'' \emph{Intelligent service robotics}, vol.~9, no.~1, pp. 1--29,
  2016.

\bibitem{hastie_stuetzle_1989}
T.~Hastie and W.~Stuetzle, ``Principal curves,'' \emph{Journal of the American
  Statistical Association}, vol.~84, no. 406, p. 502–516, 1989.

\bibitem{gorban_visualization2001}
A.~Gorban and A.~Zinovyev, ``Visualization of data by method of elastic maps
  and its applications in genomics, economics and sociology,'' 2001.

\bibitem{piecewiseSkeletonization}
B.~Kégl and A.~Krzyzak, ``Piecewise linear skeletonization using principal
  curves,'' \emph{IEEE Transactions on Pattern Analysis and Machine
  Intelligence}, 2002.

\bibitem{sumner2007embedded}
R.~W. Sumner, J.~Schmid, and M.~Pauly, ``Embedded deformation for shape
  manipulation,'' in \emph{ACM siggraph 2007 papers}, 2007, pp. 80--es.

\bibitem{whelan2016elasticfusion}
T.~Whelan, R.~F. Salas-Moreno, B.~Glocker, A.~J. Davison, and S.~Leutenegger,
  ``Elasticfusion: Real-time dense slam and light source estimation,''
  \emph{The International Journal of Robotics Research}, vol.~35, no.~14, pp.
  1697--1716, 2016.

\bibitem{douglas_peucker1973algorithms}
D.~H. Douglas and T.~K. Peucker, ``Algorithms for the reduction of the number
  of points required to represent a digitized line or its caricature,''
  \emph{Cartographica: the international journal for geographic information and
  geovisualization}, vol.~10, no.~2, pp. 112--122, 1973.

\bibitem{Khansari-Zadeh2011LASA}
S.~M. Khansari-Zadeh and A.~Billard, ``Learning stable nonlinear dynamical
  systems with gaussian mixture models,'' \emph{IEEE Transactions on Robotics},
  vol.~27, no.~5, pp. 943--957, 2011.

\bibitem{Eiter1994frechet_computing}
T.~Eiter and H.~Mannila, ``Computing discrete fr{\'e}chet distance,'' Citeseer,
  Tech. Rep., 1994.

\bibitem{ontanon2020overview}
S.~Onta{\~n}{\'o}n, ``An overview of distance and similarity functions for
  structured data,'' \emph{Artificial Intelligence Review}, vol.~53, no.~7, pp.
  5309--5351, 2020.

\bibitem{Rana2020GT_dataset}
R.~M. Asif, C.~Dephne, W.~Jacob, C.~Vivian, S.~R. Ahmadzadeh, and C.~Sonia,
  ``Benchmark for skill learning from demonstration: Impact of user experience,
  task complexity, and start configuration on performance,'' in \emph{Robotics
  and Automation ({ICRA}), {IEEE} International Conference on}.\hskip 1em plus
  0.5em minus 0.4em\relax Paris, France: {IEEE}, May 2020, pp. 7561--7567.

\bibitem{Shoemake1985slerp}
K.~Shoemake, ``Animating rotation with quaternion curves,'' in
  \emph{Proceedings of the 12th annual conference on Computer graphics and
  interactive techniques}, 1985, pp. 245--254.

\end{thebibliography}

\end{document}